\documentclass{article}

\usepackage[nonatbib, preprint]{neurips_2026}

\usepackage[utf8]{inputenc} %
\usepackage[T1]{fontenc}    %
\usepackage{hyperref}       %
\usepackage{url}            %
\usepackage{booktabs}       %
\usepackage{amsfonts}       %
\usepackage{nicefrac}       %
\usepackage{microtype}      %
\usepackage{xcolor}         %

\usepackage{amsmath}
\usepackage{amsthm}
\usepackage{amssymb}
\usepackage{marvosym}
\usepackage{graphicx}
\usepackage{xspace}
\usepackage{inconsolata}
\usepackage{float}
\usepackage{multirow}
\usepackage{adjustbox}
\usepackage{threeparttable}
\usepackage{multirow}
\usepackage{wrapfig}
\usepackage{tipa}
\usepackage{siunitx}
\usepackage{nicematrix}
\usepackage[style=numeric-comp,sorting=none,natbib=true,backend=biber,minbibnames=10,maxbibnames=10,giveninits=true,doi=false,isbn=false,url=false]{biblatex}
\DeclareFieldFormat{title}{\mkbibquote{#1}}
\DeclareFieldFormat{eprint:arxiv}{arXiv: \texttt{#1}}
\DeclareFieldFormat*{eprint}{\texttt{#1}}
\DeclareFieldFormat*{doi}{\textsc{doi:} #1}
\renewbibmacro{in:}{}

\newcommand{\possm}{\texttt{POSSM}\xspace}
\newcommand{\pogru}{\texttt{POGRU}\xspace}
\newcommand{\pomamba}{\texttt{POMAMBA}\xspace}
\newcommand{\poxlstm}{\texttt{POXLSTM}\xspace}
\newcommand{\pomambatwo}{\texttt{POMAMBA2}\xspace}
\newcommand{\possmplus}{\texttt{POSSM+}\xspace}

\newcommand{\poyo}{\texttt{POYO}\xspace}
\newcommand{\poyoplus}{\texttt{POYO+}\xspace}

\newcommand{\mojo}{\texttt{MOJO}\xspace}

\DeclareMathOperator{\resultTaken}{\dagger}

\hypersetup{
    colorlinks,
    linkcolor={blue!64!black},
    citecolor={blue!64!black},
    urlcolor={blue!64!black}
}

\usepackage[textsize=tiny]{todonotes}

\title{Leveraging unlabelled data for generalizable neural population decoding}

\author{%
Ximeng Mao$^{*,1,2,\text{\Letter}}$ \quad 
Nanda H Krishna$^{*,1,2,\text{\Letter}}$
\\
\textbf{Avery Hee-Woon Ryoo}$^{1,2}$ \quad 
\textbf{Matthew G Perich}$^{\dagger,1,2}$ \quad 
\textbf{Guillaume Lajoie}$^{\dagger,1,2,3,\text{\Letter}}$ \\
$^{1}$Mila -- Quebec AI Institute \enspace $^{2}$Université de Montréal \enspace $^{3}$Canada CIFAR AI Chair \enspace \\ $^{*}$Co-first authors \enspace $^{\dagger}$Co-senior authors\\
$^{\text{\Letter}}${\small \href{mailto:ximengmao@gmail.com}{\texttt{ximengmao}}\texttt{@gmail.com},\texttt{\{}\href{mailto:nanda.harishankar-krishna@mila.quebec}{\texttt{nanda.harishankar-krishna}},\href{mailto:guillaume.lajoie@mila.quebec}{\texttt{guillaume.lajoie}}\texttt{\}@mila.quebec}}}

\begin{document}

\maketitle

\begin{abstract}
Robust and accurate neural decoders are integral to neurotechnologies such as brain-computer interfaces and closed-loop experiments. Recent work has shown that tokenizing neural data at the spike level facilitates multi-session pretraining and delivers state-of-the-art decoding performance. However, current spike-based models are restricted to supervised learning (SL), limiting training to datasets with paired behavioural labels. To address this limitation, we introduce \mojo (\textbf{M}asked aut\textbf{O}encoder-based \textbf{JO}int training), a training framework for spike-tokenizing models that jointly leverages self-supervised learning (SSL) via masked autoencoding and SL objectives. We evaluate \mojo on three spiking datasets spanning monkey motor cortex during reaching tasks and multi-regional mouse recordings during vision and decision making tasks, demonstrating superior performance over purely SL-trained models. This improvement is especially pronounced when training with limited labelled data, particularly in few-shot finetuning, where only a small amount of labelled data from a new session is available. Incorporating SSL also yields more interpretable neuronal representations, improving performance on brain region classification and spike-statistics prediction without explicit optimization for these tasks. We further show that \mojo generalizes beyond spiking data to human electrocorticography during speech, where it continues to outperform purely SL-trained models and achieves performance comparable to neuro-foundation models (NFMs) designed specifically for continuous signals. Overall, augmenting spike-tokenizing models with SSL improves performance in label-impoverished settings and enables the use of unlabelled data across various tasks and species, while generalizing to other neural modalities. These results suggest a path towards more flexible and scalable data usage when training NFMs.
\end{abstract}

\begin{figure*}
    \centering
    \includegraphics[width=\textwidth]{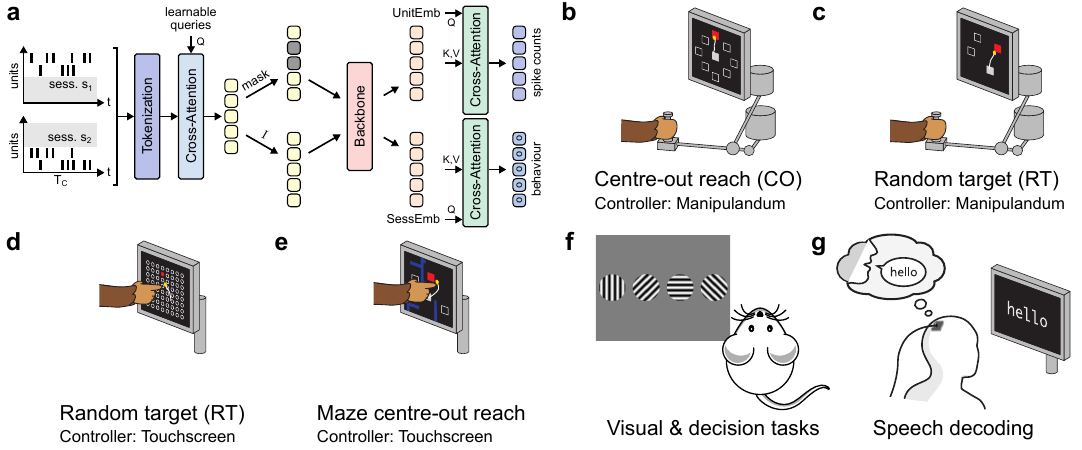}
    \caption{{\bf Model and task schematics.} (a) Schematic showing a \poyo-style model augmented with \mojo. Latent representations extracted from tokenized neural data are simultaneously used for supervised learning (SL) and self-supervised learning (SSL). The former is carried out by minimizing error in predicted behaviour while the latter is carried out by reconstructing spike counts from masked latents. (b-e) Schematics describing the monkey reaching tasks that \mojo is evaluated on. (f) Schematic of the mouse visual and decision tasks. (g) Schematic of the human speech decoding task.}
    \label{fig:mojo}
\end{figure*}

\section{Introduction}
Brain computer interfaces (BCIs) and other neurotechnologies are powered by neural decoders: models that map neural activity to some behavioural variable. Artificial neural networks (ANNs) are well-established in the literature as an effective candidate to learn these mappings, with earlier work achieving reasonable success using multilayer perceptrons (MLPs) \cite{GlaserENEURO.0506-19.2020} and recurrent networks (RNNs) \cite{cho_learning_2014,Ali_2024,sussillo_making_2016}. However, Transformers \cite{vaswani_attention_2017} and State-Space Models (SSMs) \cite{gu2024mambalineartimesequencemodeling,dao2024transformersssmsgeneralizedmodels} have recently emerged as particularly adept architectures, primarily due to their scalability as well as their strong decoding performance across a diverse set of modalities and tasks \cite{dosovitskiy2021an,jiang_large_2023,lieber_jamba_2024}. These architectures are now widely applied to varied neural data modalities \cite{zhang2023brant,wang2023brainbertselfsupervisedrepresentationlearning,azabou2023unified}, notably on invasive neural spiking data where different ways to process and tokenize input signals are under rapid development.

Among these approaches, the \poyo model family \cite{azabou2023unified,ryoo2025generalizable} -- characterized by a tokenization scheme that operates at single-spike resolution -- is particularly notable. This contrasts with the commonly-adopted approach that bins spike counts \cite{Ye_2021_NDT,zhang2025neuralencodingdecodingscale}, which imposes a rigid input formulation and requires additional stitching layers when training across different recording sessions. The flexibility of \poyo-style spike tokens facilitates large-scale pretraining across multiple datasets, enabling efficient finetuning and high decoding accuracy across tasks.

Despite several landmark results on spike datasets, the potential of these \poyo models is still limited by a major inflexibility: their exclusive reliance on supervised learning (SL) \cite{Goodfellow-et-al-2016}. This limits their pretraining to neural datasets with paired behavioural labels (e.g., common applications include motor tasks in monkeys with neuronal spikes mapped to 2D arm velocities). This approach remains starkly in contrast to the development of foundation models such as GPT \cite{Brown2020GPT} and BERT \cite{devlin-etal-2019-bert} in the language domain, which leverage vast amounts of unlabelled text data to learn rich representations via self-supervised learning (SSL) \cite{simCLR_2020,Assran2023SelfSupervisedLF}. There exists a plethora of diverse and unannotated neural data collected from multiple neuroscience laboratories, highlighting a promising avenue for \textit{significantly} expanding the size and diversity of pretraining data for models like \poyo. Exploiting this requires a pretraining scheme that can (1) extract meaningful information about the inherent structure of this neural data and (2) accommodate heterogeneous data formats within a unified pipeline. Although recent works have explored forecasting or input-masking-based SSL objectives for \poyo-based models, those methods are designed for continuous calcium traces \cite{duan2026poco,willeke2026omnimouse}, and it remains unclear how to design an effective scheme for sequences of spike tokens.

With these goals in mind, we introduce \texttt{MOJO} (\textbf{M}asked Aut\textbf{O}encoder-based \textbf{JO}int Pretraining),
a joint pretraining framework in which SSL and SL objectives are optimized simultaneously. 
\mojo addresses the challenges of applying SSL on spike tokens by leveraging a masked autoencoder \cite{he_CVPR_MAE} on the encoded latents. Given the sparse and irregular nature of neural spikes, this avoids the need to distinguish between non-existent and masked spike tokens at the input level, and enables seamless integration with existing supervised \poyo family backbones.
A key feature of \mojo is its joint optimization of SSL and SL objectives, which encourages synergy between the representations learned from neural and behavioural data (yielding, for instance, task-informed neuron representations).

We evaluated \mojo on both backbones (Transformers and SSMs) used in the \poyo family and three intracortical spike datasets involving monkey reaching, mouse vision and decision tasks.
\mojo was further applied to a human electrocorticography (ECoG) dataset on speech articulation, to test its generalizability beyond spikes. In all these settings, \mojo consistently outperforms purely supervised methods. Incorporating SSL further enables more efficient learning in label-impoverished regimes and yields more interpretable neuronal representations. 
Moreover, we demonstrate \mojo's scaling behaviour with respect to heterogeneous data sources, showcasing positive transfer from joint cross-species training on both decoding and analysis tasks.
Our contributions are as follows:
\begin{itemize}
    \item We propose \mojo, a joint SSL-SL framework that can leverage unlabelled data in spike-tokenizing models via masked autoencoding.%
    \item We evaluate \mojo on neural datasets with varying species, modalities, tasks, and brain regions, and achieve efficient finetuning and improved decoding performance over purely SL-trained models with significant improvements even if up to 90\% of the data is unlabelled. 
    \item Our learned unit embeddings exhibit high discriminability to meta-features of neural units (e.g., source brain area and firing rate distributions), despite not being explicitly optimized for these tasks or having explicit access to such information during training.
    \item \mojo scales across heterogeneous data sources and demonstrates cross-species transfer of learned representations, through a model pretrained on monkey reaching and subsequently trained on both monkey reaching and mouse vision datasets.
\end{itemize}
\section{Methods}
As shown in Figure \ref{fig:mojo}, \mojo maintains different pathways for SSL and SL objectives. In this section, we describe the key components of \mojo along with its SSL and SL pathways. For brevity, we describe only the model components essential to \mojo and leave detailed descriptions to Appendix \ref{app:sec:modelArchi}. 
\subsection{Spike Tokenization and Input Cross-Attention} \label{sec:spiketokenization}
The spike tokenization scheme \citep{azabou2023unified} assigns each individual spike a token, represented as a tuple consisting of two components: the neural unit it came from and the exact time it occurred.
Note that no further meta-information regarding the neural unit  is encoded beyond an arbitrary integer ID.

\mojo employs a \possm-style encoder \cite{ryoo2025generalizable}, where the cross-attention is computed separately on each contiguous {\it time chunk} in the data sample. Considering an input sequence $\mathbf{X}_t$ of all spike tokens from time chunk $t$ and a latent query $\mathbf{c}^{(t)}_{i}$ from learnable vector $i$, 
the encoder output is calculated via scaled dot-product attention \cite{vaswani_attention_2017} to produce a latent token $\mathbf{z}^{(t)}_{i}$.

It is important to note that the \possm-style encoder is different from that of \poyo, where the input cross-attention is performed on the entire sequence (i.e., $\mathbf{X}$ rather than $\mathbf{X}_t$). In principle, \mojo could simultaneously maintain both encoders to support separate pathways. However, this formulation may introduce unnecessary redundancy and is therefore not enabled by default throughout the paper.
Please see Appendix \ref{app:sec:pathway} for experiments comparing the two formulations.
\subsection{Backbones}
\mojo requires a sequence model to serve as its backbone. We consider the two main architectures from existing spike-tokenizing models: attention-based \cite{azabou2023unified} and recurrence-based \cite{ryoo2025generalizable}.

\paragraph{Attention-based.}
The attention backbone is composed of stacked self-attention blocks operating over the entire latent sequence, where hidden state $\mathbf{h}^{(t)}_{l,i}$, from latent query $i$ at time $t$ from layer $l$, is calculated as 
$\mathbf{h}^{(t)}_{l,i}=\mathrm{softmax}(\mathbf{q}^{(t)}_l\mathbf{K}^\top/\sqrt{D})\mathbf{V}$,
where $\mathbf{q}^{(t)}_l$ is projected from $\mathbf{h}^{(t)}_{l-1,i}$ and $\mathbf{h}^{(t)}_{0,i} = \mathbf{z}^{(t)}_i$,
\paragraph{Recurrence-based.}
The recurrent backbone is a stack of SSM blocks where the hidden state is updated following the transition function
$\mathbf{h}^{(t)}_l = f_{\text{SSM}}(\mathbf{h}^{(t)}_{l-1}, \mathbf{h}^{(t-1)}_l)$, where $\mathbf{h}^{(t)}_0$ is the concatenation of $N_c$ latent tokens in the same time chunk, that is $\mathbf{h}^{(t)}_0 = [\mathbf{z}^{(t)}_1,...,\mathbf{z}^{(t)}_{N_c}]$.
\subsection{Masked Autoencoder}
In the SSL pathway, temporal masking is applied onto the latent outputs from the input cross-attention, inspired by the masking strategy in \citep{wav2vec}. This is implemented using a mask indicator $m^t$ which is a Bernoulli variable with $p=0.5$, with which all the latents at $t$ are transformed via:
\begin{align*}
\mathbf{z}^{(t)}_{i} = (1 - m^{t})\mathbf{z}^{(t)}_{i}+ m^{t}\mathbf{e}_{\mathrm{mask}}, \forall i \in [1,...,N_c],
\end{align*}
where $\mathbf{e}_{\mathrm{mask}}$ is a learnable mask token. The combination of temporal masking and local cross-attention within each time chunk ensures that all spiking information from the masked intervals is effectively removed from the SSL pathway.
\subsection{Output Cross-Attention}
For the SL pathway, we adopt the multi-task setting \cite{azabou2025multisession}, where output queries for different tasks cross-attend to the same latent sequence. The outputs are then projected via separate linear readouts for each of the tasks.
The SSL pathway predicts spike rates with a unit query $\mathbf{u}$, similar to the input spike token.
Note that the format of unit query is conceptually similar to that in \citet{duan2026poco} but for spikes. The unit query is then decoded via the same cross-attention operations as the SL pathway, but with the constraint that $\mathbf{u}$ can only attend to the latent tokens situated within the same time chunk. 
\subsection{Training Objectives}
The \mojo training objective, $L_{\mathrm{MOJO}} = \alpha_{\mathrm{SSL}}L_{\mathrm{SSL}} + \alpha_{\mathrm{SL}} L_{\mathrm{SL}}$, is a weighted sum of the losses from the two pathways,
with $\alpha_{\mathrm{SSL}}$ and $\alpha_{\mathrm{SL}}$ as coefficients.
$L_{\mathrm{SSL}}$ is the Poisson negative log-likelihood of the predicted spike-rates, and $L_{\mathrm{SL}}$ 
is a weighted sum over individual supervised tasks in the multi-task setting.
Empirically, we found that setting both pathway coefficients to $1$ yields strong performance across all experiments, even if labelled and unlabelled training set sizes differ significantly.

This joint objective encourages the model to learn the underlying neural population dynamics (SSL pathway), while simultaneously ensuring that the learned representations remain task-relevant (SL pathway), all achieved jointly through gradient-based learning.
\subsection{Pathway Integration}\label{sec:integration}
While the SL and SSL pathways have unique advantages, maintaining completely separate sets of parameters is costly. We therefore integrate the two pathways by 1) sharing the output of the input cross-attention and 2) sharing the backbone parameters. This introduces only a modest increase in parameters for the SSL pathway, due to one additional output cross-attention module which accounts for approximately 10\% and 2\% of the total parameters, when using \possm and \poyo backbones, respectively. Overall, this pathway integration enables a parameter-efficient training process.

\section{Experiments}
\label{sec:experiments}
\subsection{Monkey Reaching Tasks}
\label{sec:monkeyexperiment}
The first set of experiments focuses on reaching tasks performed by non-human primates, using a collection of five public monkey datasets from various labs. These datasets encompass center-out (CO), random target (RT), and maze navigation tasks. We decode two-dimensional hand velocities from neural spiking activity recorded by electrode arrays implanted in primary motor, dorsal premotor, and primary somatosensory cortices. 
Following this, we evaluate \mojo using the same pipeline and dataset splits described in \citet{ryoo2025generalizable}. 

We focused on \mojo with three backbones: \pogru (\possm with GRU \citep{cho_learning_2014}), \pomamba (\possm with Mamba \citep{gu2024mambalineartimesequencemodeling}) and \poyo.
We used 20 ms time bins for all SL pathways, consistent with SSL and a sequence length of 1 s for each training sample. Additionally, we adopted the causal evaluation strategy presented in \citet{ryoo2025generalizable}, to emphasize the capability of real-time decoding. We employed the two finetuning strategies from \citet{azabou2023unified} when adapting the pretrained model to previously unseen data sessions: unit identification (UI) and full finetuning (FT), where UI updates only the unit and session embeddings (see Appendix \ref{app:sec:finetuning} for more details).
\begin{table}[t]
\centering
\vspace{5mm}
\caption{{\bf Behavioural decoding results on monkey reaching tasks.} Values are mean $R^2$ $\pm$ SD over sessions. Best models are in boldface (1st) and underlined (2nd). *: $p < 0.05$ on a paired t-test over sessions vs. the best non-MOJO baseline. $^{\resultTaken}$Results reproduced from \citet{ryoo2025generalizable}.}
\vspace{0.1in}
\label{tab:monkey20ms}
{\renewcommand{\arraystretch}{1.2}
\adjustbox{max width=0.65\textwidth}{\begin{NiceTabular}{l>{\enspace}c<{\enspace}>{\enspace}cc}
\toprule
   & \textit{Same animal, other days} & \multicolumn{2}{c}{\textit{New animal}} \\
 \cmidrule(lr){2-2} \cmidrule(lr){3-4} 
  Method & C – CO 2010 (5) & T – CO (6) & T – RT (6) \\ \midrule
  MLP$^{\resultTaken}$ & {0.5842 $\pm$ 0.2052} & 0.7940 $\pm$ 0.0341 & 0.6082 $\pm$ 0.3014 \\
  GRU$^{\resultTaken}$ & 0.7742 $\pm$ 0.0964 & 0.8389 $\pm$ 0.0248 & 0.7414 $\pm$ 0.0426 \\
 \midrule
  \poyo (UI) & 0.7759 $\pm$ 0.1003 & 0.8123 $\pm$ 0.0419 & 0.7011 $\pm$ 0.0975 \\ 
  \pomamba (UI) & 0.7283 $\pm$ 0.1138 & 0.8574 $\pm$ 0.0225 & 0.7283 $\pm$ 0.0846 \\
  \pogru (UI) & 0.7632 $\pm$ 0.1013 & 0.8587 $\pm$ 0.0216 & 0.7331 $\pm$ 0.0775 \\
  \mojo-\poyo (UI) & 0.7846 $\pm$ 0.0897 & 0.8387 $\pm$ 0.0251 & 0.7470 $\pm$ 0.0617 \\
  \mojo-\pomamba (UI) & \underline{0.7937} $\pm$ 0.0735* & \underline{0.8753} $\pm$ 0.0160* & \textbf{0.7591} $\pm$ 0.0588* \\
  \mojo-\pogru (UI) & \textbf{0.7949} $\pm$ 0.0892* & \textbf{0.8772} $\pm$ 0.0189* & \underline{0.7570} $\pm$ 0.0693*\\
 \midrule
  NDT-2$^{\resultTaken}$ & 0.7846 $\pm$ 0.1167 & 0.7173 $\pm$ 0.0443 & 0.6323 $\pm$ 0.1339 \\
  NDT-3 & 0.7524 $\pm$ 0.1322 & 0.8576 $\pm$ 0.0313 & 0.7066 $\pm$ 0.0980 \\
  NEDS & 0.5968 $\pm$ 0.0760 & 0.7635 $\pm$ 0.0758 & 0.6121 $\pm$ 0.0918\\
  \poyo (FT) & \underline{0.8244} $\pm$ 0.0753 & 0.8817 $\pm$ 0.0352 & 0.7624 $\pm$ 0.0815 \\
  \pomamba (FT) & 0.8142 $\pm$ 0.0763 & 0.8949 $\pm$ 0.0152 & 0.7580 $\pm$ 0.0745 \\
  \pogru (FT) & 0.8126 $\pm$ 0.0892 & 0.8936 $\pm$ 0.0212 & 0.7575 $\pm$ 0.0875 \\ 
  \mojo-\poyo (FT) & \textbf{0.8438} $\pm$ 0.0888* & \textbf{0.9131} $\pm$ 0.0177* & \textbf{0.7964} $\pm$ 0.0663* \\
  \mojo-\pomamba (FT) & 0.8153 $\pm$ 0.0925 & 0.9043 $\pm$ 0.0182* & 0.7675 $\pm$ 0.0806 \\
  \mojo-\pogru (FT) & 0.8222 $\pm$ 0.0950 & \underline{0.9103} $\pm$ 0.0168* & \underline{0.7776} $\pm$ 0.0710* \\
 \bottomrule
\end{NiceTabular}}}
\end{table}

\paragraph{Results.}
In Table \ref{tab:monkey20ms}, we show the generalization performance of pretrained \mojo for decoding when transferred to new recording sessions and previously unseen animals. As baselines, we compare against purely supervised methods based on spike tokenization (\pogru, \pomamba and \poyo), as well as other SSL-based decoders that use binning (NDT-2 \cite{ye_neural_2023}, NDT-3 \cite{ye2025a}, and NEDS \cite{zhang2025neuralencodingdecodingscale}).
We find that \mojo outperforms these state-of-the-art methods on all datasets considered. All models trained with \mojo achieved improved decoding performance when finetuned on each held-out session, relative to their purely supervised counterparts under the same finetuning strategy (FT or UI). Notably, results from the more challenging sessions (unseen monkey T-RT) showed that performing UI alone with \mojo-\possm ($<18$K trainable parameters) was on par with fully finetuned (FT) supervised \possm ($>7.6$M trainable parameters). Complete results on monkey reaching tasks are shown in Appendix Table \ref{app:tab:monkey20ms} including other \possm backbone variants and single-session results. 
\paragraph{Few-shot Finetuning.}
\begin{wrapfigure}[18]{r}{0.5\textwidth}
    \centering
    \includegraphics[width=\linewidth]{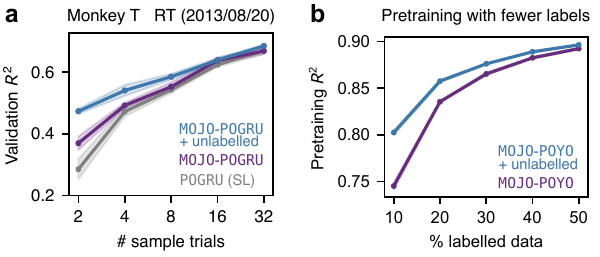}
    \caption{{\bf Leveraging unlabelled data for finetuning and pretraining.} (a) \mojo improves few-shot finetuning performance over standard SL and leverages additional unlabelled data to improve performance further. (b) \mojo can improve decoding performance by exploiting all available unlabelled data even when little labelled data is available.}
    \label{fig:monkeyfewshot}
\end{wrapfigure}

\mojo also enables robust and efficient few-shot adaptation. Figure \ref{fig:monkeyfewshot}a shows decoding performance  for \mojo-\pogru and \pogru when finetuned on only a few trials of labelled data, including new sessions from an unseen animal. \mojo consistently outperforms the supervised baseline in the low-data regime, with the largest gains observed when labelled data is scarce. Further, we show that \mojo can leverage additional unlabelled data during finetuning: In the monkey T-RT session shown, incorporating up to 32 trials of unlabelled data allows the model to achieve over $60\%$ and $75\%$ of the fully supervised performance using only two and four labelled calibration trials, respectively.
\paragraph{Finetuning with Random Behaviours.}
\mojo was further evaluated on a more challenging finetuning setting in which all spiking data are retained, but behavioural labels are restricted to a sparse sub-phase (\texttt{RANDOM}) that lies outside the cued reaching periods in each CO session. Any movements in this period are self-initiated and likely akin to fidgeting. \mojo achieved an average $R^2$ of $0.5237$ when finetuned on new sessions of a seen animal, and exceeds $0.1598$ $R^2$ in 3 out of the 6 new sessions from an unseen animal. These results reveal that a pretrained \mojo can infer behaviours in structured tasks by extrapolating from the same behavioural variables observed during other time periods. Please see Appendix \ref{app:sec:phase} for detailed descriptions of the phases in CO sessions, and more results on transferring across phases. 
\paragraph{Unlabelled Data During Pretraining.}
In addition to finetuning, \mojo can effectively leverage unlabelled data during pretraining. We conducted a series of experiments on the Perich et al. dataset when varying the percentage of labelled data available during pretraining from 10\% to 50\% (Figure \ref{fig:monkeyfewshot}b). In the limited label regime, \mojo's ability to exploit the full set of unlabelled data enables superior pretraining performance, while preserving discriminable unit embeddings. 
Please see Appendix \ref{app:sec:unlabelledPretrain} for results on brain region classification with the pretrained embeddings and comparison when finetuning using these models.
\paragraph{Unit Embedding Analyses.}
\setlength{\intextsep}{0.0pt}
\begin{wraptable}[21]{r}{6.5cm}
\centering
\caption{{\bf Visual stimuli classification results on mouse vision tasks.} Values are mean accuracy $\pm$ SD over sessions. 
Best performing models are in boldface (1st) and underlined (2nd).}
\vspace{4mm}
\resizebox{6.5cm}{!}{%
{\renewcommand{\arraystretch}{1.2}
\begin{NiceTabular}{l>{\enspace}cc<{\enspace}}
\toprule
   & \multicolumn{2}{c}{\textit{Stimulus Set 1}}\\
 \cmidrule(lr){2-3}
  Method & NS (4) Acc. (\%) & DG (4) Acc. (\%) \\ \midrule
 MLP & 83.17 $\pm$ 4.92 & 90.18 $\pm$ 2.03 \\
 \midrule
 \pogru (UI) & 88.87 $\pm$ 2.55 & 99.17 $\pm$ 0.76 \\
 \poyo (UI) & 88.17 $\pm$ 3.17 & 93.54 $\pm$ 2.06 \\ 
  \mojo-\pogru (UI) & 92.54 $\pm$ 3.41 & 99.53 $\pm$ 0.20 \\
  \mojo-\poyo (UI) & 91.53 $\pm$ 1.99 & \underline{99.74} $\pm$ 0.31 \\
  \mojo-\poyo(J) (UI) & \underline{93.15} $\pm$ 2.39 & 99.43 $\pm$ 1.15 \\
  \mojo-\poyo-L(J) (UI) & \textbf{94.18} $\pm$ 2.00 & \textbf{99.84} $\pm$ 0.20 \\
 \midrule
    NEDS & 93.96 $\pm$ 2.38 & 99.20 $\pm$ 0.57 \\
  \pogru (FT) & 91.09 $\pm$ 5.64 & 99.48 $\pm$ 0.62 \\
  \poyo (FT) & 91.03 $\pm$ 4.07 & 93.39 $\pm$ 4.50 \\ 
  \mojo-\pogru (FT) & 94.47 $\pm$ 2.83 & 99.22 $\pm$ 0.91 \\
  \mojo-\poyo (FT) & 94.12 $\pm$ 2.79 & \textbf{100.00} $\pm$ 0.00 \\
  \mojo-\poyo(J) (FT) & \underline{94.62} $\pm$ 2.74 & 99.69 $\pm$ 0.64 \\
  \mojo-\poyo-L(J) (FT) & \textbf{95.48} $\pm$ 2.05 & \underline{99.74} $\pm$ 0.52 \\
 \bottomrule
\end{NiceTabular}}}
\medskip
\label{tab:mouse20ms}
\end{wraptable}
We analyzed the unit embeddings learned during pretraining using a linear probe on their associated metadata (brain regions and subject ID) via logistic regression. The probing was conducted at both the single-dataset (using units from the same dataset) and multi-dataset (using units pooled across all datasets) level. As shown in Appendix Table \ref{app:tab:monkeyregionclf}, \mojo yields interpretable neuronal embeddings with high discriminability using only a linear classifier, despite having no access to electrode, region, or subject metadata during pretraining. 
\subsection{Mouse Vision Tasks}
In the second set of experiments, we evaluate the performance of \mojo on decoding visual stimuli from multi-region mouse recordings. We used the Allen visual coding dataset \cite{allenVisualCodingDataset,allenVisualCodingDataset_whitePaper}, where mouse spiking activity was recorded by Neuropixels probes from a wide range of brain regions, including multiple areas of the visual cortex, thalamus, and hippocampus. There are 58 data sessions in total, spanning the four transgenic lines of the recorded animals and two stimulus sets. We held out eight sessions for evaluation (one per transgenic line and per stimulus set) and used the remaining 50 for pretraining. For this experiment, we focused on three visual decoding tasks spanning natural and artificial stimuli: natural scene (NS) classification, drifting grating (DG) orientations, and drifting grating temporal frequency (TF) classification.

In this experiment, we considered \mojo-\poyo and \mojo-\pogru. All models were allowed to look at the entire trial before making a prediction. For \possm, we placed the behaviour timestamps at the end of each trial. 
As with the monkey experiment, we used 1 s sequence lengths and 20 ms time chunks.
\paragraph{Results.}
Decoding performance of \mojo and baseline methods are summarized in Table \ref{tab:mouse20ms}, including NS and DG accuracies from evaluation sessions in stimulus set 1. \mojo demonstrates improved accuracies on both DG and NS over pretrained supervised counterparts, with nearly perfect performance on DG in both UI and FT.
Complete results on the mouse vision tasks are shown in Appendix Table \ref{app:tab:mouse20ms} including all the tasks as well as single-session results. 
\paragraph{Joint Monkey-Mouse Pretraining.}
Next, we explore joint pretraining on both the monkey reaching and mouse vision datasets with \mojo, through a curriculum learning strategy in which \mojo is first pretrained on the aforementioned monkey datasets then jointly on both datasets. For the jointly trained model (\mojo-\poyo(J)), we were able to observe a positive transfer via faster convergence on the mouse NS tasks (see Appendix Table \ref{app:tab:valjoint}) while maintaining consistent performance on the monkey reaching. In addition, \mojo-\poyo(J) was also found to produce more intepretable unit embeddings and achieve improved finetuning performance on NS (note that other tasks are already near perfect). Moreover, we note that the performance of joint pretraining can be further improved with a larger model (\mojo-\poyo-L(J)). Appendix \ref{app:sec:transfer} includes more results on joint pretraining.

\begin{figure*}
    \centering
    \includegraphics[width=\textwidth]{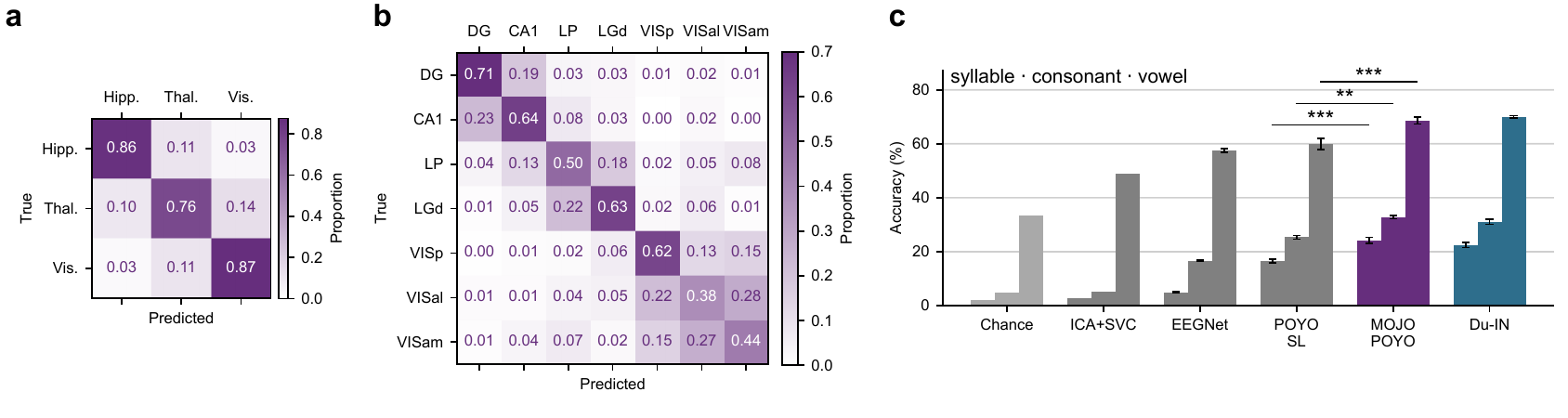}
    \vspace{-1em}
    \caption{{\bf Mouse brain region classification and human speech decoding results.} (a-b) Confusion matrices for (a) 3-class and (b) 7-class brain region classification of neurons in mice from the Allen visual coding dataset. (c) Classification accuracy of syllables, consonants, and vowels for speech decoding from human electrocorticography. For (a-c), we report the mean accuracy over sessions averaged across 5 seeds. (**,***): $p < 0.01, 0.001$ resp. for a paired t-test across all sessions. \mojo vs. Du-IN is n.s. for the t-test.}
    \label{fig:brcecog}
\end{figure*}

\paragraph{Unit Embedding Analyses.}
Trained unit embeddings from all 50 pretraining sessions were used to linearly probe their associated brain regions. When training the linear classifier, we did not differentiate between brain regions from different animals; thus, the learned decision boundaries reflect population-level structure rather than subject-specific effects. We included 7 brain regions in the analyses: cornu ammonis 1 (CA1), dentate gyrus (DG), lateral posterior nucleus (LP), lateral geniculate nucleus (LGd), primary visual cortex (VISp), anterolateral visual area (VISal) and anteromedial visual area (VISam). These regions were well represented among the spike-sorted units and overall covered the main brain areas included in the data set. Figure \ref{fig:brcecog}a-b shows the confusion matrices of the learned linear classifiers from \mojo-\poyo-L(J) on the 3 broader brain areas -- visual, hippocampus and thalamus -- and the complete 7 brain regions.
Despite inter-subject variability, the probability mass is primarily concentrated along the diagonal, and analysis of the errors suggested similarities among specific regions based on their location and function.

Beyond categorical region identity, we further asked whether unit embeddings encode fine-grained single-neuron properties (Figure~\ref{fig:unitmetadata}, Appendix~\ref{app:sec:unitmetadata}). On the same 50 pretraining sessions, we linearly regressed 18 spike-statistic features per neuron \citep{schneider2022} -- including moments of the inter-spike interval (ISI) distribution, firing rate, local-variability indices (CV, LV), gamma-distribution fits, and band-limited spike-train PSDs -- from the unit embeddings of our models. \mojo unit embeddings predict every target substantially better than pure-SL \poyo (e.g., $R^2 = 0.61$ vs.\ $0.33$ for log median ISI, $0.88$ vs.\ $0.54$ for log mean firing rate, and $0.18$ vs.\ $0.00$ for CV). The embedding geometry also reflects probe topology: across 1.23M pairs of units recorded on the same Neuropixels shank, the cosine similarity of \mojo embeddings is more strongly anti-correlated with electrode distance than \poyo (Spearman $r$ between $-0.25$ and $-0.29$ for \mojo variants vs.\ $-0.12$ for \poyo, all pairwise differences highly significant via paired bootstrap, $p<10^{-300}$). Notably, when restricted to cross-area pairs this relation is non-monotonic for \mojo~-- with a pronounced bump at $\sim$4\,mm probe distance dominated by visual-cortex $\leftrightarrow$ visual thalamic (LGd, LP) and midbrain (APN) pairs -- indicating that \mojo brings functionally related but anatomically distant units close together in embedding space, a structure largely absent in pure-SL \poyo.

These analyses verify that \mojo can learn interpretable unit embeddings across multiple brain regions, and also verify that SSL allows unit embeddings to better encode neuronal firing properties, purely through end-to-end training. Note that this analysis is different from that of previous work \cite{azabou2025multisession}, where such classification was done on session-averaged latent outputs of the input cross-attention. We found that this capability scales directly with the amount of neural data, with classification accuracy improving monotonically as more unlabelled data is added to the training pipeline of \mojo-\poyo. Adding unlabelled data from the same visual coding dataset resulted in a $5.5\%$ improvement on 7 region classification, and incorporating additional monkey data resulted in further improvements of $2.5 \%$. Lastly, as with decoding performance, a larger joint model can further improve the accuracy by $ 3.5 \%$. Please see Appendix \ref{app:sec:mouseprobing} and Table \ref{app:tab:mousebrainregionclf} for more details and baseline comparisons.
\subsection{Mouse Decision Tasks}
We conducted a third set of experiments on IBL Reproducible Electrophysiology datasets \cite{iblReproduceEphysDataset}, where mouse multi-region Neuropixel recordings were collected in multiple labs from one repeated site, during visually-driven decision-making tasks. From a total of 84 sessions, we followed prior work \citep{zhang2025neuralencodingdecodingscale} and performed four behavioural decoding tasks. We trained on 74 sessions and held-out 10 pre-determined sessions for finetuning. This dataset was previously evaluated in a non-causal fashion \cite{zhang2025neuralencodingdecodingscale}; accordingly, we used the \poyo backbone for \mojo.
In contrast to \citet{zhang2025neuralencodingdecodingscale}, for \mojo and \poyo we removed both the trial-alignment constraints and the neuron-exclusion steps based on firing rate from the training data, while ensuring that the labelled behaviour data were only from the pre-determined intervals. 
Each pre-determined interval was 2 s long, and \mojo was trained with a sequence length of 1.6 s and time chunk length of 20 ms.
\paragraph{Results.}
\setlength{\intextsep}{0pt}
\begin{wraptable}[11]{r}{6.0cm}
\centering
\caption{{\bf Mouse decision tasks results.} Values for choice and block are mean balanced accuracy and for wheel and whisker $R^{2}$, all with $\pm$ SD over sessions. *: $p < 0.05$ on a paired t-test over sessions vs POYO.}
\vspace{3mm}
\scalebox{0.7}{
{\renewcommand{\arraystretch}{1.2}
\begin{NiceTabular}{lccc}
\toprule
  Task & NEDS & \poyo & \mojo-\poyo \\
  \midrule
  Choice & 0.820 $\pm$ 0.10 & 0.882 $\pm$ 0.09\enspace & 0.913 $\pm$ 0.07* \\
  Block & 0.779 $\pm$ 0.10 & 0.848 $\pm$ 0.05\enspace & 0.864 $\pm$ 0.04\enspace \\
  Wheel & 0.515 $\pm$ 0.09 & 0.630 $\pm$ 0.10\enspace & 0.665 $\pm$ 0.09* \\
  Whisker & 0.475 $\pm$ 0.11 & 0.550 $\pm$ 0.11\enspace & 0.581 $\pm$ 0.09* \\
  Avg. & 0.647 $\pm$ 0.09 & 0.727 $\pm$ 0.07\enspace & 0.756 $\pm$ 0.05* \\
 \bottomrule
\end{NiceTabular}}}
\medskip
\label{tab:ibl}
\end{wraptable}
As shown in Table \ref{tab:ibl}, \mojo demonstrates advantages over pure SL models in this task, showcasing the importance of joint SSL-SL training. Due to known split changes and re-sorting of the spiking data, the decoding results are not directly comparable with those reported in \citet{zhang2025neuralencodingdecodingscale}. Thus, we reran NEDS using the original implementation on the new sessions to report its results (see Appendix \ref{app:sec:iblneds}).
\subsection{Human Speech Tasks}
Finally, we tested the viability of \mojo beyond spiking data using a human ECoG speech articulation dataset \cite{bouchard_ecog,bouchard2019ecogdata} recorded from four participants (with 30 total sessions) speaking consonant-vowel syllables. ECoG signals were collected with a high-density 256-channel array implanted in ventral sensorimotor cortex during epilepsy treatment while participants read commonly-used syllables in American English. We extracted the high-gamma band and downsampled to 200 Hz following the original study. This dataset presents a multi-task classification problem, where neural decoders were trained to predict the corresponding vowel, consonant, and full syllable pairs from ECoG activity. For each syllable trial, we used a 1.2 s time window centered on the consonant-vowel transition time provided in the dataset, from 550 ms before to 650 ms after. Each training sample is a 1 s interval drawn from one trial, and we applied \mojo directly to each data sample by further dividing the interval into 20 ms patches, each with a patch size of 4. For these experiments, we used \poyo backbone with value embeddings \citep{azabou2025multisession} to be compatible with the continuous signal format.
\paragraph{Results.}
\mojo was trained on all sessions across every participant, with test accuracies shown in Figure \ref{fig:brcecog}c. Compared to purely SL trained \poyo, \mojo shows consistent improvements across all tasks, especially on more challenging one ($47 \%$ improvement for classifying syllables). 
\mojo's performance is also comparable to a recent EEG foundation model Du-IN \cite{zheng2024duin}, showing that \mojo remains competitive on this dataset despite being designed for spike data. 
Overall, these results suggest that \mojo can generalize across neural modalities beyond spikes, highlighting its potential for future multi-modal training.
\section{Related Works}
\paragraph{Traditional Neural Decoders.} Historically, neural decoding for continuous tasks such as cursor velocity decoding in BCIs was accomplished using simple statistical methods such as the Kalman filter \citep{Kalman1960,wu2002neural,koyama2010comparison,Willett2019}. Subsequent works also leveraged simple artificial neural network models such as MLPs \citep{GlaserENEURO.0506-19.2020} and RNNs \citep{Sussillo2012} for continuous decoding tasks with great success in single-session settings. With the advent of more sophisticated deep learning methods such as the variational RNNs and sequential VAEs \citep{chung2016recurrentlatentvariablemodel}, LFADS \citep{lfads} was proposed to extract temporally-varying latent variables from high-dimensional neural spiking data. Such models have also been used to perform neural decoding tasks in real-time, as shown in recent work \citep{Ali_2024}. A key drawback of such approaches, however, is their limited ability to scale to large-scale multi-session datasets of neural recordings, and the difficulty in adapting single-session models to new recording sessions or subjects \citep{sussillo_making_2016}.

\paragraph{Pretraining for Neural Decoding.}
With instrumental advances in large-scale deep learning models for sequential data \citep{vaswani_attention_2017,gu_efficiently_2021,dao2024transformersssmsgeneralizedmodels} and the availability of large-scale public datasets of neural recordings, a slew of recent works have proposed neural decoding models that leverage Transformers or State-Space Models combined with large-scale pretraining to achieve state-of-the-art decoding performance and generalizability to new sessions, tasks, subjects, and even species. State-of-the-art pretraining-based decoders for neural spikes can be largely classified into two categories based on the way they ingest neural data. The first category, i.e., NDT-style models such as NDT-2 \citep{ye_neural_2023}, NDT-MtM \citep{Zhang_2024_arXiv}, NDT-3 \citep{ye2025a}, and NEDS \citep{zhang2025neuralencodingdecodingscale}, process binned or patched spike counts to decode behaviour. A key feature of all these models is the incorporation of self-supervised learning objectives -- NDT-2 and NDT-MtM carry out self-supervised pretraining followed by downstream finetuning, NEDS is jointly trained on neural encoding and decoding with several masking schemes, and NDT-3 autoregressively predicts neural data and behaviour in a flat sequence. 

The second category, i.e., \poyo-style models, process each neural spike as a separate token, allowing these models to rely on fine-grained temporal information for prediction and naturally facilitating large-scale pretraining and efficient finetuning \citep{azabou2023unified,ryoo2025generalizable}. These spike-based models have achieved state-of-the-art performance on several neural decoding tasks and demonstrated successful transfer of decoders to new sessions, tasks, subjects, and even species \citep{ryoo2025generalizable}. However, they have thus far been trained only with supervised learning objectives and cannot leverage unlabelled data to improve their performance -- a gap that this work aims to fill. Beyond spikes, \poyo-style architectures have been successfully applied to calcium traces for decoding \citep{azabou2025multisession} and forecasting \citep{duan2026poco}, considering fine-grained temporal patches as tokens. Most closely related to our work, \citet{willeke2026omnimouse} proposed OmniMouse for calcium traces using a \poyo-style architecture with both SSL and SL. However, it uses meta-data–integrated tokenization and 1D convolutions (unsuitable for spikes), and focuses on learning across multiple modalities (calcium, behaviour, video) with a NEDS-like architecture.
\section{Discussion}
Our results demonstrate that our joint SSL-SL training framework, \mojo, enables improved decoding performance, more efficient finetuning, and tractable neuronal analyses across diverse datasets, compared to purely supervised \poyo-family models. These gains are achieved with minimal parameter overhead (e.g., on an additional $231$K parameters in a $9.88$M parameter model with \poyo) since most of the parameters are shared between the SSL and SL objectives via the proposed pathway integration strategy. We see significant downstream improvement even if up to 90\% of the pretraining data is unlabelled. 
This places \mojo in a unique position to harness vast amounts of unlabelled spiking data while maintaining the scalability and flexibility of \poyo-style tokenization. Finally, we demonstrate that \mojo can perform cross-species joint pretraining, with positive transfer from monkey reaching to mouse visual processing, and applicable beyond spiking data, paving the way for training unified and scalable neuro-foundation models.

Being a joint SSL-SL model, empirically we find \mojo to be data-hungry, as it needs a sufficient amount of data to learn proper neural dynamics. Otherwise, a weakly-learned SSL objective could even be detrimental to the SL performance. The latent masking scheme of \mojo, despite offering flexibility when integrating with SL pathway, is limited to temporal masking, and an explicit objective to infer spatially (unseen neurons or brain regions) could further boost the SSL performance. Moreover, as highly structured and interpretable as the learned unit embeddings during pretraining (as evidenced by accurate brain-region classification), when finetuning to a new session, the unit embeddings need to be re-learned from scratch. This is a limitation inherited from the \poyo family of models but becomes especially wasteful with SSL. Lastly, while \mojo is easily adaptable to ECoG, it still requires additional linear layer for value embeddings during tokenization.
This modality-specific design constrains the diversity of datasets that can be jointly incorporated during pretraining.

To address these limitations, future work will explore joint multi-modal training \cite{zhang2025neuralencodingdecodingscale,ye2025a,willeke2026omnimouse}. As datasets from different modalities could represent diverse neural functions, this is an effective direction to enrich the pretraining data, and prediction across modalities can encourage more comprehensive understandings of the underling neural dynamics. Amortized methods \cite{arora2025know} and discrete codes \cite{jiang_large_2023} are promising future directions for encouraging reuse of the learned model for new unit embeddings.

\printbibliography

@book{Goodfellow-et-al-2016,
    title={Deep Learning},
    author={Ian Goodfellow and Yoshua Bengio and Aaron Courville},
    publisher={MIT Press},
    note={\url{http://www.deeplearningbook.org}},
    year={2016}
}

@dataset{odoherty2017nonhuman,
	title        = {Nonhuman Primate Reaching with Multichannel Sensorimotor Cortex Electrophysiology},
	author       = {O'Doherty, Joseph E. and Cardoso, Mariana M. B. and Makin, Joseph G. and Sabes, Philip N.},
	year         = 2020,
	note         = {Zenodo: \texttt{10.5281/zenodo.3854034}}
}

@inproceedings{wang2023brainbertselfsupervisedrepresentationlearning,
	title        = {Brain{BERT}: Self-supervised representation learning for intracranial recordings},
	author       = {Christopher Wang and Vighnesh Subramaniam and Adam Uri Yaari and Gabriel Kreiman and Boris Katz and Ignacio Cases and Andrei Barbu},
	year         = 2023,
	booktitle    = {The Eleventh International Conference on Learning Representations},
}

@misc{gu2024mambalineartimesequencemodeling,
	title        = {Mamba: Linear-Time Sequence Modeling with Selective State Spaces},
	author       = {Albert Gu and Tri Dao},
	year         = 2024,
	eprint       = {2312.00752},
	archiveprefix = {arXiv},
	primaryclass = {cs.LG}
}

@article{WillettHandwriting,
	title        = {High-performance brain-to-text communication via handwriting},
	author       = {Willett,  Francis R. and Avansino,  Donald T. and Hochberg,  Leigh R. and Henderson,  Jaimie M. and Shenoy,  Krishna V.},
	year         = 2021,
	journal      = {Nature},
	volume       = 593,
	number       = 7858,
	pages        = {249–254},
}

@article{GlaserENEURO.0506-19.2020,
	title        = {Machine Learning for Neural Decoding},
	author       = {Glaser, Joshua I. and Benjamin, Ari S. and Chowdhury, Raeed H. and Perich, Matthew G. and Miller, Lee E. and Kording, Konrad P.},
	year         = 2020,
	journal      = {eNeuro},
	volume       = 7,
	number       = 4,
}

@inproceedings{azabou2023unified,
	title        = {A Unified, Scalable Framework for Neural Population Decoding},
	author       = {Azabou, Mehdi and Arora, Vinam and Ganesh, Venkataramana and Mao, Ximeng and Nachimuthu, Santosh and Mendelson, Michael and Richards, Blake and Perich, Matthew and Lajoie, Guillaume and Dyer, Eva},
	year         = 2023,
	booktitle    = {Advances in Neural Information Processing Systems},
	volume       = 36,
	pages        = {44937--44956},
}

@article{schneider2022,
title = {Transcriptomic cell type structures in vivo neuronal activity across multiple timescales},
journal = {Cell Reports},
volume = {42},
number = {4},
pages = {112318},
year = {2023},
issn = {2211-1247},
doi = {https://doi.org/10.1016/j.celrep.2023.112318},
url = {https://www.sciencedirect.com/science/article/pii/S2211124723003297},
author = {Aidan Schneider and Mehdi Azabou and Louis McDougall-Vigier and David F. Parks and Sahara Ensley and Kiran Bhaskaran-Nair and Tomasz Nowakowski and Eva L. Dyer and Keith B. Hengen}
}

@inproceedings{azabou2025multisession,
title={Multi-session, multi-task neural decoding from distinct cell-types and brain regions},
author={Mehdi Azabou and Krystal Xuejing Pan and Vinam Arora and Ian Jarratt Knight and Eva L Dyer and Blake Aaron Richards},
booktitle={The Thirteenth International Conference on Learning Representations},
year={2025},
}

@inproceedings{ryoo2025generalizable,
title={Generalizable, real-time neural decoding with hybrid state-space models},
author={Avery Hee-Woon Ryoo and Nanda H Krishna and Ximeng Mao and Mehdi Azabou and Eva L Dyer and Matthew G Perich and Guillaume Lajoie},
booktitle={Advances in Neural Information Processing Systems},
year={2025},
}

@article{Ye_2021_NDT,
	title        = {Representation learning for neural population activity with {Neural} {Data} {Transformers}},
	author       = {Ye, Joel and Pandarinath, Chethan},
	year         = 2021,
	journal      = {Neurons, Behavior, Data analysis, and Theory},
	volume       = 5,
	number       = 3,
	pages        = {1--18},
}

@inproceedings{dosovitskiy2021an,
	title        = {An Image is Worth 16x16 Words: Transformers for Image Recognition at Scale},
	author       = {Alexey Dosovitskiy and Lucas Beyer and Alexander Kolesnikov and Dirk Weissenborn and Xiaohua Zhai and Thomas Unterthiner and Mostafa Dehghani and Matthias Minderer and Georg Heigold and Sylvain Gelly and Jakob Uszkoreit and Neil Houlsby},
	year         = 2021,
	booktitle    = {International Conference on Learning Representations},
}

@inproceedings{jaegle2022perceiver,
	title        = {Perceiver {IO}: A General Architecture for Structured Inputs \& Outputs},
	author       = {Andrew Jaegle and Sebastian Borgeaud and Jean-Baptiste Alayrac and Carl Doersch and Catalin Ionescu and David Ding and Skanda Koppula and Daniel Zoran and Andrew Brock and Evan Shelhamer and Olivier J Henaff and Matthew Botvinick and Andrew Zisserman and Oriol Vinyals and Joao Carreira},
	year         = 2022,
	booktitle    = {International Conference on Learning Representations},
}

@article{su2023roformerenhancedtransformerrotary,
	title        = {RoFormer: Enhanced transformer with Rotary Position Embedding},
	author       = {Jianlin Su and Murtadha Ahmed and Yu Lu and Shengfeng Pan and Wen Bo and Yunfeng Liu},
	year         = 2024,
	journal      = {Neurocomputing},
	volume       = 568,
	pages        = 127063,
}

@misc{perich_miller_2018_dataset,
	title        = {Long-term recordings of motor and premotor cortical spiking activity during reaching in monkeys},
	author       = {Perich, Matthew G. and Miller, Lee E. and Azabou, Mehdi and Dyer, Eva L.},
	year         = 2025,
}

@article{Kalman1960,
	title        = {A New Approach to Linear Filtering and Prediction Problems},
	author       = {Kalman, R. E.},
	year         = 1960,
	journal      = {Journal of Basic Engineering},
	volume       = 82,
	number       = 1,
	pages        = {35--45},
}

@inproceedings{wu2002neural,
	title        = {Neural Decoding of Cursor Motion Using a Kalman Filter},
	author       = {Wu, W and Black, M. and Gao, Y. and Serruya, M. and Shaikhouni, A. and Donoghue, J. and Bienenstock, Elie},
	year         = 2002,
	booktitle    = {Advances in Neural Information Processing Systems},
	volume       = 15,
}

@article{koyama2010comparison,
	title        = {Comparison of brain–computer interface decoding algorithms in open-loop and closed-loop control},
	author       = {Koyama,  Shinsuke and Chase,  Steven M. and Whitford,  Andrew S. and Velliste,  Meel and Schwartz,  Andrew B. and Kass,  Robert E.},
	year         = 2010,
	journal      = {Journal of Computational Neuroscience},
	volume       = 29,
	number       = {1–2},
	pages        = {73--87},
}

@article{Willett2019,
	title        = {Principled BCI Decoder Design and Parameter Selection Using a Feedback Control Model},
	author       = {Willett,  Francis R. and Young,  Daniel R. and Murphy,  Brian A. and Memberg,  William D. and Blabe,  Christine H. and Pandarinath,  Chethan and Stavisky,  Sergey D. and Rezaii,  Paymon and Saab,  Jad and Walter,  Benjamin L. and Sweet,  Jennifer A. and Miller,  Jonathan P. and Henderson,  Jaimie M. and Shenoy,  Krishna V. and Simeral,  John D. and Jarosiewicz,  Beata and Hochberg,  Leigh R. and Kirsch,  Robert F. and Bolu Ajiboye,  A.},
	year         = 2019,
	journal      = {Scientific Reports},
	volume       = 9,
	number       = 1,
}

@article{Churchland2012,
	title        = {Neural population dynamics during reaching},
	author       = {Churchland,  Mark M. and Cunningham,  John P. and Kaufman,  Matthew T. and Foster,  Justin D. and Nuyujukian,  Paul and Ryu,  Stephen I. and Shenoy,  Krishna V.},
	year         = 2012,
	journal      = {Nature},
	volume       = 487,
	number       = 7405,
	pages        = {51–56},
}

@article{Ali_2024,
	title        = {BRAND: a platform for closed-loop experiments with deep network models},
	author       = {Ali, Yahia H and Bodkin, Kevin and Rigotti-Thompson, Mattia and Patel, Kushant and Card, Nicholas S and Bhaduri, Bareesh and Nason-Tomaszewski, Samuel R and Mifsud, Domenick M and Hou, Xianda and Nicolas, Claire and Allcroft, Shane and Hochberg, Leigh R and Au Yong, Nicholas and Stavisky, Sergey D and Miller, Lee E and Brandman, David M and Pandarinath, Chethan},
	year         = 2024,
	journal      = {Journal of Neural Engineering},
	volume       = 21,
	number       = 2,
	pages        = {026046},
}

@inproceedings{vaswani_attention_2017,
	title        = {Attention is All you Need},
	author       = {Vaswani, Ashish and Shazeer, Noam and Parmar, Niki and Uszkoreit, Jakob and Jones, Llion and Gomez, Aidan N and Kaiser, \L ukasz and Polosukhin, Illia},
	year         = 2017,
	booktitle    = {Advances in Neural Information Processing Systems},
	volume       = 30,
}

@inproceedings{simCLR_2020,
author = {Chen, Ting and Kornblith, Simon and Norouzi, Mohammad and Hinton, Geoffrey},
title = {A simple framework for contrastive learning of visual representations},
year = {2020},
booktitle = {Proceedings of the 37th International Conference on Machine Learning},
articleno = {149},
numpages = {11},
}

@article{lfads,
	title        = {Inferring single-trial neural population dynamics using sequential auto-encoders},
	author       = {Pandarinath,  Chethan and O’Shea,  Daniel J. and Collins,  Jasmine and Jozefowicz,  Rafal and Stavisky,  Sergey D. and Kao,  Jonathan C. and Trautmann,  Eric M. and Kaufman,  Matthew T. and Ryu,  Stephen I. and Hochberg,  Leigh R. and Henderson,  Jaimie M. and Shenoy,  Krishna V. and Abbott,  L. F. and Sussillo,  David},
	year         = 2018,
	journal      = {Nature Methods},
	volume       = 15,
	number       = 10,
	pages        = {805–815},
}

@inproceedings{ye_neural_2023,
	title        = {Neural Data Transformer 2: Multi-context Pretraining for Neural Spiking Activity},
	author       = {Ye, Joel and Collinger, Jennifer and Wehbe, Leila and Gaunt, Robert},
	year         = 2023,
	booktitle    = {Advances in Neural Information Processing Systems},
	volume       = 36,
	pages        = {80352--80374},
}

@article{Flint_2012,
	title        = {Accurate decoding of reaching movements from field potentials in the absence of spikes},
	author       = {Flint, Robert D and Lindberg, Eric W and Jordan, Luke R and Miller, Lee E and Slutzky, Marc W},
	year         = 2012,
	journal      = {Journal of Neural Engineering},
	volume       = 9,
	number       = 4,
	pages        = {046006},
}

@article{sussillo_making_2016,
	title        = {Making brain–machine interfaces robust to future neural variability},
	author       = {Sussillo,  David and Stavisky,  Sergey D. and Kao,  Jonathan C. and Ryu,  Stephen I. and Shenoy,  Krishna V.},
	year         = 2016,
	journal      = {Nature Communications},
	volume       = 7,
	number       = 1,
}

@inproceedings{lieber_jamba_2024,
	title        = {Jamba: Hybrid Transformer-Mamba Language Models},
	author       = {Barak Lenz and Opher Lieber and Alan Arazi and Amir Bergman and Avshalom Manevich and Barak Peleg and Ben Aviram and Chen Almagor and Clara Fridman and Dan Padnos and Daniel Gissin and Daniel Jannai and Dor Muhlgay and Dor Zimberg and Edden M. Gerber and Elad Dolev and Eran Krakovsky and Erez Safahi and Erez Schwartz and Gal Cohen and Gal Shachaf and Haim Rozenblum and Hofit Bata and Ido Blass and Inbal Magar and Itay Dalmedigos and Jhonathan Osin and Julie Fadlon and Maria Rozman and Matan Danos and Michael Gokhman and Mor Zusman and Naama Gidron and Nir Ratner and Noam Gat and Noam Rozen and Oded Fried and Ohad Leshno and Omer Antverg and Omri Abend and Or Dagan and Orit Cohavi and Raz Alon and Ro'i Belson and Roi Cohen and Rom Gilad and Roman Glozman and Shahar Lev and Shai Shalev-Shwartz and Shaked Haim Meirom and Tal Delbari and Tal Ness and Tomer Asida and Tom Ben Gal and Tom Braude and Uriya Pumerantz and Josh Cohen and Yonatan Belinkov and Yuval Globerson and Yuval Peleg Levy and Yoav Shoham},
	year         = 2025,
	booktitle    = {The Thirteenth International Conference on Learning Representations},
}

@inproceedings{zhang2025neuralencodingdecodingscale,
	title        = {Neural Encoding and Decoding at Scale},
	author       = {Zhang, Yizi and Wang, Yanchen and Azabou, Mehdi and Andre, Alexandre and Wang, Zixuan and Lyu, Hanrui and Laboratory, International Brain and Dyer, Eva L and Paninski, Liam and Hurwitz, Cole Lincoln},
	year         = 2025,
	booktitle    = {Proceedings of the 42nd International Conference on Machine Learning},
	series       = {Proceedings of Machine Learning Research},
	volume       = 267,
	pages        = {76175--76192},
}

@inproceedings{Zhang_2024_arXiv,
	title        = {Towards a "Universal Translator" for Neural Dynamics at Single-Cell, Single-Spike Resolution},
	author       = {Zhang, Yizi and Wang, Yanchen and Jim\'{e}nez-Benet\'{o}, Donato M. and Wang, Zixuan and Azabou, Mehdi and Richards, Blake and Tung, Renee and Winter, Olivier and Laboratory, The International Brain and Dyer, Eva and Paninski, Liam and Hurwitz, Cole},
	year         = 2024,
	booktitle    = {Advances in Neural Information Processing Systems},
	volume       = 37,
	pages        = {80495--80521},
}

@inproceedings{jiang_large_2023,
	title        = {Large Brain Model for Learning Generic Representations with Tremendous {EEG} Data in {BCI}},
	author       = {Weibang Jiang and Liming Zhao and Bao-liang Lu},
	year         = 2024,
	booktitle    = {The Twelfth International Conference on Learning Representations},
}

@inproceedings{gu_efficiently_2021,
	title        = {Efficiently Modeling Long Sequences with Structured State Spaces},
	author       = {Albert Gu and Karan Goel and Christopher Re},
	year         = 2022,
	booktitle    = {International Conference on Learning Representations},
}

@inproceedings{cho_learning_2014,
	title        = {Learning Phrase Representations using {RNN} Encoder{--}Decoder for Statistical Machine Translation},
	author       = {Cho, Kyunghyun  and van Merri{\"e}nboer, Bart  and Gulcehre, Caglar  and Bahdanau, Dzmitry  and Bougares, Fethi  and Schwenk, Holger  and Bengio, Yoshua},
	year         = 2014,
	booktitle    = {Proceedings of the 2014 Conference on Empirical Methods in Natural Language Processing ({EMNLP})},
	address      = {Doha, Qatar},
	pages        = {1724--1734},
}

@inproceedings{dao2024transformersssmsgeneralizedmodels,
	title        = {Transformers are {SSM}s: Generalized Models and Efficient Algorithms Through Structured State Space Duality},
	author       = {Dao, Tri and Gu, Albert},
	year         = 2024,
	booktitle    = {Proceedings of the 41st International Conference on Machine Learning},
	volume       = 235,
	pages        = {10041--10071},
}

@inproceedings{PeiYe2021NeuralLatents,
	title        = {Neural Latents Benchmark ‘21: Evaluating latent variable models of neural population activity},
	author       = {Pei, Felix and Ye, Joel and Zoltowski, David and Wu, Anqi and Chowdhury, Raeed and Sohn, Hansem and O\textquotesingle Doherty, Joseph and Shenoy, Krishna V and Kaufman, Matthew and Churchland, Mark and Jazayeri, Mehrdad and Miller, Lee and Pillow, Jonathan and Park, Il Memming and Dyer, Eva and Pandarinath, Chethan},
	year         = 2021,
	booktitle    = {Proceedings of the Neural Information Processing Systems Track on Datasets and Benchmarks},
	volume       = 1,
}

@misc{lamb,
      title={Large Batch Optimization for Deep Learning: Training BERT in 76 minutes}, 
      author={Yang You and Jing Li and Sashank Reddi and Jonathan Hseu and Sanjiv Kumar and Srinadh Bhojanapalli and Xiaodan Song and James Demmel and Kurt Keutzer and Cho-Jui Hsieh},
      year={2020},
      eprint={1904.00962},
      archivePrefix={arXiv},
      primaryClass={cs.LG}, 
}

@inproceedings{adamw,
title={Decoupled Weight Decay Regularization},
author={Ilya Loshchilov and Frank Hutter},
booktitle={International Conference on Learning Representations},
year={2019},
}

@inproceedings{ye2025a,
title={A Generalist Intracortical Motor Decoder},
author={Joel Ye and Fabio Rizzoglio and Xuan Ma and Adam Smoulder and Hongwei Mao and Gary H Blumenthal and William Hockeimer and Nicolas Guazzelli Kunigk and Dalton D. Moore and Patrick J. Marino and Raeed H. Chowdhury and J. Patrick Mayo and Aaron Batista and Steven Chase and Michael L Boninger and Charles M. Greenspon and Andrew B. Schwartz and Nicholas G. Hatsopoulos and Lee E. Miller and Kristofer Bouchard and Jennifer L Collinger and Leila Wehbe and Robert Gaunt},
booktitle={Advances in Neural Information Processing Systems},
year={2025},
}

@inproceedings{Brown2020GPT,
 author = {Brown, Tom and Mann, Benjamin and Ryder, Nick and Subbiah, Melanie and Kaplan, Jared D and Dhariwal, Prafulla and Neelakantan, Arvind and Shyam, Pranav and Sastry, Girish and Askell, Amanda and Agarwal, Sandhini and Herbert-Voss, Ariel and Krueger, Gretchen and Henighan, Tom and Child, Rewon and Ramesh, Aditya and Ziegler, Daniel and Wu, Jeffrey and Winter, Clemens and Hesse, Chris and Chen, Mark and Sigler, Eric and Litwin, Mateusz and Gray, Scott and Chess, Benjamin and Clark, Jack and Berner, Christopher and McCandlish, Sam and Radford, Alec and Sutskever, Ilya and Amodei, Dario},
 booktitle = {Advances in Neural Information Processing Systems},
 pages = {1877--1901},
 title = {Language Models are Few-Shot Learners},
 volume = {33},
 year = {2020}
}

@inproceedings{devlin-etal-2019-bert,
    title = "{BERT}: Pre-training of Deep Bidirectional Transformers for Language Understanding",
    author = "Devlin, Jacob  and
      Chang, Ming-Wei  and
      Lee, Kenton  and
      Toutanova, Kristina",
    booktitle = "Proceedings of the 2019 Conference of the North {A}merican Chapter of the Association for Computational Linguistics: Human Language Technologies, Volume 1 (Long and Short Papers)",
    month = jun,
    year = "2019",
    address = "Minneapolis, Minnesota",
    pages = "4171--4186",
}

@article {allenVisualCodingDataset,
article_type = {journal},
title = {Sharing neurophysiology data from the Allen Brain Observatory},
author = {de Vries, Saskia EJ and Siegle, Joshua H and Koch, Christof},
editor = {Meister, Markus and Gold, Joshua I and Meister, Markus and Chen, Jerry L},
volume = 12,
year = 2023,
month = jul,
pub_date = {2023-07-11},
pages = {e85550},
citation = {eLife 2023;12:e85550},
journal = {eLife},
issn = {2050-084X},
publisher = {eLife Sciences Publications, Ltd},
}

@article{iblReproduceEphysDataset, 
 title={Reproducibility of in vivo electrophysiological measurements in mice}, 
 url={http://dx.doi.org/10.7554/eLife.100840.2}, 
 DOI={10.7554/elife.100840.2}, 
 journal = {eLife},
 publisher={eLife Sciences Publications, Ltd}, 
 author={International Brain Laboratory IBL and Banga, Kush and Benson, Julius and Bhagat, Jai and Biderman, Dan and Birman, Daniel and Bonacchi, Niccolò and Bruijns, Sebastian A and Buchanan, Kelly and Campbell, Robert AA and Carandini, Matteo and Chapuis, Gaëlle A and Churchland, Anne K and Davatolhagh, M Felicia and Lee, Hyun Dong and Faulkner, Mayo and Gerçek, Berk and Hu, Fei and Huntenburg, Julia and Hurwitz, Cole and Khanal, Anup and Krasniak, Christopher and Langfield, Christopher and Lau, Petrina and Mackenzie, Nancy and Meijer, Guido T and Miska, Nathaniel J and Mohammadi, Zeinab and Noel, Jean-Paul and Paninski, Liam and Pan-Vazquez, Alejandro and Rossant, Cyrille and Roth, Noam and Schartner, Michael and Socha, Karolina and Steinmetz, Nicholas A and Svoboda, Karel and Taheri, Marsa and Urai, Anne E and Wang, Shuqi and Wells, Miles and West, Steven J and Whiteway, Matthew R and Winter, Olivier and Witten, Ilana B and Zhang, Yizi}, 
 year={2025}, 
 month=mar,
 }

@article{DBLP:journals/corr/abs-1807-03748,
  author       = {A{\"{a}}ron van den Oord and
                  Yazhe Li and
                  Oriol Vinyals},
  title        = {Representation Learning with Contrastive Predictive Coding},
  journal      = {CoRR},
  volume       = {abs/1807.03748},
  year         = {2018},
  url          = {http://arxiv.org/abs/1807.03748},
  eprinttype   = {arXiv},
  eprint       = {1807.03748},
  bibsource    = {dblp computer science bibliography, https://dblp.org}
}

@article{zhang2024torcheeg,
    title = {{TorchEEGEMO}: A deep learning toolbox towards {EEG}-based emotion recognition},
    journal = {Expert Systems with Applications},
    pages = {123550},
    year = {2024},
    issn = {0957-4174},
    author = {Zhi Zhang and Sheng-hua Zhong and Yan Liu}
}

@article{Sussillo2012,
  title = {A recurrent neural network for closed-loop intracortical brain–machine interface decoders},
  volume = {9},
  ISSN = {1741-2552},
  number = {2},
  journal = {Journal of Neural Engineering},
  publisher = {IOP Publishing},
  author = {Sussillo,  David and Nuyujukian,  Paul and Fan,  Joline M and Kao,  Jonathan C and Stavisky,  Sergey D and Ryu,  Stephen and Shenoy,  Krishna},
  year = {2012},
  month = mar,
  pages = {026027}
}

@misc{chung2016recurrentlatentvariablemodel,
      title={A Recurrent Latent Variable Model for Sequential Data}, 
      author={Junyoung Chung and Kyle Kastner and Laurent Dinh and Kratarth Goel and Aaron Courville and Yoshua Bengio},
      year={2016},
      eprint={1506.02216},
      archivePrefix={arXiv},
      primaryClass={cs.LG}, 
}

@article {allenVisualCodingDataset_whitePaper,
article_type = {journal},
title = {Survey of spiking in the mouse visual system reveals functional hierarchy},
author = {Siegle, J. H. and Jia, X. and Durand, S. and Gale, S. and Bennett, C. and Graddis, N. and Heller, G. and Ramirez, T. K. and Choi, H. and Luviano, J. A. and Groblewski, P. A. and Ahmed, R. and Arkhipov, A. and Bernard, A. and Billeh, Y. N. and Brown, D. and Buice, M. A. and Cain, N. and Caldejon, S. and Casal, L. and others},
volume = 592,
year = 2021,
pages = {86--92},
journal = {Nature},
issn = {2050-084X},
}

@inproceedings{zhang2023brant,
title={Brant: Foundation Model for Intracranial Neural Signal},
author={Daoze Zhang and Zhizhang Yuan and Yang Yang and Junru Chen and Jingjing Wang and Yafeng Li},
booktitle={Advances in Neural Information Processing Systems},
year={2023},
}

@inproceedings{he_CVPR_MAE,
  author={He, Kaiming and Chen, Xinlei and Xie, Saining and Li, Yanghao and Dollár, Piotr and Girshick, Ross},
  booktitle={2022 IEEE/CVF Conference on Computer Vision and Pattern Recognition (CVPR)}, 
  title={Masked Autoencoders Are Scalable Vision Learners}, 
  year={2022},
  volume={},
  number={},
  pages={15979-15988},
}

@inproceedings{Assran2023SelfSupervisedLF,
  title={Self-Supervised Learning from Images with a Joint-Embedding Predictive Architecture},
  author={Mahmoud Assran and Quentin Duval and Ishan Misra and Piotr Bojanowski and Pascal Vincent and Michael G. Rabbat and Yann LeCun and Nicolas Ballas},
  journal={2023 IEEE/CVF Conference on Computer Vision and Pattern Recognition (CVPR)},
  year={2023},
  pages={15619-15629},
}

@inproceedings{arora2025know,
title={Know Thyself by Knowing Others: Learning Neuron Identity from Population Context},
author={Vinam Arora and Divyansha Lachi and Ian Jarratt Knight and Mehdi Azabou and Blake Aaron Richards and Cole Lincoln Hurwitz and Josh Siegle and Eva L Dyer},
booktitle={Advances in Neural Information Processing Systems},
year={2025},
}

@inproceedings{bouchard_ecog,
    author = {Bouchard, K. E. and Mesgarani, N. and Johnson, K. and Chang, E. F.},
    title = {Functional organization of human sensorimotor cortex for speech articulation},
    booktitle = {Nature},
    year = 2013,
    pages = {327--332},
    volumn = {495},
    number = {7441},
}

@inproceedings{beck:24xlstm,
  title = {xLSTM: Extended Long Short-Term Memory}, 
  author = {Maximilian Beck and Korbinian Pöppel and Markus Spanring and Andreas Auer and Oleksandra Prudnikova and Michael Kopp and Günter Klambauer and Johannes Brandstetter and Sepp Hochreiter},
  booktitle = {Advances in Neural Information Processing Systems},
  year = {2024}, 
}

@inproceedings{
duan2026poco,
title={{POCO}: Scalable Neural Forecasting through Population Conditioning},
author={Yu Duan and Hamza Tahir Chaudhry and Misha B. Ahrens and Christopher D Harvey and Matthew G Perich and Karl Deisseroth and Kanaka Rajan},
booktitle={The Thirty-ninth Annual Conference on Neural Information Processing Systems},
year={2026},
url={https://openreview.net/forum?id=Xma2G1ak3H}
}

@inproceedings{
willeke2026omnimouse,
title={OmniMouse: Scaling properties of multi-modal, multi-task Brain Models on 150B Neural Tokens},
author={Konstantin Friedrich Willeke and Polina Turishcheva and Alex Gilbert and Goirik Chakrabarty and Hasan Atakan Bedel and Paul G. Fahey and Yongrong Qiu and Marissa A. Weis and Michaela Vystr{\v{c}}ilov{\'a} and Taliah Muhammad and Lydia Ntanavara and Rachel E Froebe and Kayla Ponder and Zheng Huan Tan and Emin Orhan and Erick Cobos and Sophia Sanborn and Katrin Franke and Fabian H. Sinz and Alexander S. Ecker and Andreas S. Tolias},
booktitle={The Fourteenth International Conference on Learning Representations},
year={2026},
url={https://openreview.net/forum?id=mEw4lhAn0F}
}

@inproceedings{wav2vec,
author = {Baevski, Alexei and Zhou, Henry and Mohamed, Abdelrahman and Auli, Michael},
title = {wav2vec 2.0: a framework for self-supervised learning of speech representations},
year = {2020},
booktitle = {Proceedings of the 34th International Conference on Neural Information Processing Systems},
articleno = {1044},
numpages = {12},
location = {Vancouver, BC, Canada},
}

@inproceedings{
zheng2024duin,
title={Du-{IN}: Discrete units-guided mask modeling for decoding speech from Intracranial Neural signals},
author={Hui Zheng and Haiteng Wang and Weibang Jiang and Zhongtao Chen and Li He and Peiyang Lin and Penghu Wei and Guoguang Zhao and Yunzhe Liu},
booktitle={The Thirty-eighth Annual Conference on Neural Information Processing Systems},
year={2024},
url={https://openreview.net/forum?id=uyLtEFnpQP}
}

@misc{bouchard2019ecogdata,
  author       = {Bouchard, Kristofer E. and Chang, Edward F.},
  title        = {Human ECoG speaking consonant-vowel syllables},
  year         = {2019},
  publisher    = {figshare},
  url          = {https://figshare.com/collections/Human_ECoG_speaking_consonant-vowel_syllables/4617263}
}

\newpage
\appendix
\section*{Supplementary Material}
\section{Additional Details on Datasets}
\subsection{Monkey Reaching Tasks}
\label{app:sec:monkeyDataset}
For experiments on the monkey reaching tasks, we adopted the same pretraining datasets and splits as those described in \citet{ryoo2025generalizable}. For the sake of completeness, we list the details of these datasets in Table \ref{tab:datasets}. During pretraining, 19 sessions were held-out from the model and served for evaluation, including 2 sessions from monkey C (20161013 \& 20161021) and all 12 sessions from monkey T from the \citet{perich_miller_2018_dataset} dataset, and all 5 sessions from \citet{Flint_2012}.
\begin{table}[h]
\centering
\medskip
\caption{{\bf Summary of datasets used on monkey motor tasks.} Dataset statistics reproduced from \citet{ryoo2025generalizable}.}
{\renewcommand{\arraystretch}{1.2}
\adjustbox{max width=\textwidth}{\begin{NiceTabular}{lccS[table-format=1.0]S[table-format=2.0]S[table-format=5.0]cr}
\toprule
 {{Dataset}} & \multicolumn{1}{c}{{Regions}} & \multicolumn{1}{c}{{Tasks}} & \multicolumn{1}{c}{{\#~Indiv.}} & \multicolumn{1}{c}{{\#~Sess.}} & \multicolumn{1}{c}{{\#~Units}} & \multicolumn{1}{c}{{\#~Spikes}} & \multicolumn{1}{c}{{\#~Bhvr.}} \\ \midrule
 \citet{perich_miller_2018_dataset}                & M1, PMd                     & CO, RT                    & 3                                  & 107                              & 10245     & 104.6M & 14.0M                   \\
 \citet{odoherty2017nonhuman}                 & M1, S1                      & RT                        & 2                                  & 44                              & 14899 & 87.8M    & 10.4M                   \\
 \citet{Churchland2012}            & M1, PMd                          & CO                        & 2                                  & 10                               & 1911      & 739M & 85.0M                   \\
 NLB Maze \cite{PeiYe2021NeuralLatents}                     & M1, PMd                          & Maze                      & 1                                  & 1                               & 182      & 3.64M & 6.81M                   \\
 \citet{Flint_2012}   & M1                          & CO                        & 1                                  & 5                               & 957     & 7.88M & 318K                    \\ \bottomrule
\end{NiceTabular}}}
\label{tab:datasets}
\end{table}
\subsection{Mouse Vision Tasks}
The Allen visual coding Neuropixel dataset \cite{allenVisualCodingDataset,allenVisualCodingDataset_whitePaper} includes 58 sessions of mouse recordings during vision tasks. Thirty sessions of the recordings were made in wildtype mice, with the rest of sessions recorded in 3 transgenic lines: Pvalb (8), Sst (12), and Vip (8). 
Animals in each session were shown one of two stimulus sets: ``Brain Observatory 1.1'' and ``Functional Connectivity''. Drifting gratings stimuli were presented in both stimulus sets (2 s per trial), while natural scenes stimuli were presented only in stimulus set 1 (250 ms per trial) with no inter-trial gray period. A total of three multi-class stimulus classification tasks are possible with this data, namely, classifying drifting grating orientations (8-class), drifting grating temporal frequencies (5-class), and natural scenes (119-class). Note that the drifting gratings stimulus in stimulus set 2 only had 4 potential orientations and 1 temporal frequency. 
\subsection{Mouse Decision Tasks}
IBL Reproducible Electrophysiology dataset \cite{iblReproduceEphysDataset} include 84 sessions of mouse recordings during decision-making tasks of head-fixed mice on location of visual gratings. The data sessions are from 10 labs but inserting probes targeting the same locations. Following existing work on this dataset \cite{zhang2025neuralencodingdecodingscale}, we include four behaviour tasks: classification tasks -- decoding choice (2 classes on left or right) and block prior (3 classes on prior probability of visual gratings: 20/80$\%$ on the right, 80/20$\%$ on the left, and 50/50$\%$; and regression tasks -- decoding wheel velocity and whisker motion energy. The objective is to decode the behaviour variables in each pre-determined trial-aligned 2s interval surrounding \texttt{stimOn\_time} (500 ms before and 1500 ms after), and for continuous variables like wheel and whisker, they are resampled to 50 Hz. We use pre-determined splits across sessions provided in this dataset. We also would like to note that our dataset preparation was different from \citet{zhang2025neuralencodingdecodingscale} in the following three aspects: Firstly, we didn't enforce trial-alignment during training, and the training intervals were extended to the union of 1) the pre-determined intervals and 2) the full trial durations including additional $200 ms$ before and after; Secondly, we do not perform neuron exclusion according to firing rate, and we included spikes from all the recorded neurons; Thirdly, for wheel and whisker, instead of normalization based on minimal and maximal values, we only normalized whisker based on a constant factor $2.5$, and the value is selected so that signal amplitude is similar between the two. 
\subsection{Human Speech Tasks}
Human ECoG speech dataset \cite{bouchard_ecog,bouchard2019ecogdata} includes 30 session of ECoG activities across four participants, reading consonant-vowel syllable pair appearing on a screen. Among the four participants, in total 57 syllable types are present in the data sessions, each with similar amount of repetitions. These syllable pairs are constituted by 3 vowels and 19 consonants. However, there are several outliers existed in the datasets, mostly in sessions of participant EC9. During preprocessing, we gave each syllable type (including the outlier) an unique index, but only keep the non-outlier trials as targets. The data splitting in this dataset was done temporally, where the last $20 \%$ of the trials are used for testing. Note that overlapping in trials is present in this dataset, hence to prevent leaking in neural activity to testing set, we dropped any trials overlapping with any of the test trials. Then the remaining trials were applied a 1:9 ratio, for validation and training set, respectively.
\section{Model Architectures}
\label{app:sec:modelArchi}
\subsection{Spike Tokenization}
The spike tokenization scheme \citep{azabou2023unified} assigns each individual spike as a token, represented as a tuple consisting of two components: the neural unit it came from and the exact time it occurred.  
\begin{align*}
\mathbf{x} = \left(\mathrm{UnitEmb}(I_{\mathrm{spike}}), T_{\mathrm{spike}}\right),
\end{align*}
where $\mathrm{UnitEmb}(\cdot)$ is a look-up table mapping every neural unit ID to a $D$-dimensional embedding and $T_{\mathrm{spike}}$ is a rotary positional embedding (RoPE) \cite{su2023roformerenhancedtransformerrotary} of the timestamp. In accordance with the literature, we adopt the term ``neural unit'' as a \textit{general} descriptor based on the dataset, encompassing individual neurons, multi-unit activity from a single electrode, as well as individual channels. In the case of ECoG, the signals from each channel within each millisecond patch are embedded via an additional linear layer, and concatenated with the unit embedding \cite{azabou2025multisession}.
\begin{align*}
\mathbf{x} = \left(\big[\mathrm{UnitEmb}(I_{\mathrm{ch.}}), \mathbf{W}_s(S_{ch., T_{\mathrm{patch}}}) \big], T_{\mathrm{patch}}\right),
\end{align*}
Note that no further meta information (like brain regions) regarding the neuron or recording channel beyond an arbitrary integer ID is encoded in this scheme.

Following \poyo, in addition to the spike tokens, we feed special tokens to the encoder indicating all the units being recorded within each data sequence. This is done with the intention of providing information on the units that did not fire. Unlike \poyo, where two sets of special tokens are utilized as delimiters and given timestamps of either the start or the end of the sequence, we place one set of tokens in every time chunk and provide them a position embedding based on the timestamp in the middle of the chunk.
\subsection{Input Cross-Attention}
Like \poyo and its variants, we use a cross-attention module to encode a variable-length input token sequence into a fixed-length latent token sequence of size $N_z$. $N_z$ is a hyperparameter, commonly chosen to be $N_z \ll \mathrm{InputLength}$, thereby effectively compressing information. We can view latent token sequence as a concatenation of $T_c$ groups of $N_c$ tokens (hence $N_z=N_c*T_c$), where the groups are distributed temporally within the training samples, and $T_c$ denotes the number of {\it time chunks}.

\mojo employ a \possm-style \cite{ryoo2025generalizable} encoder, where the cross-attention is computed separately on each contiguous time chunk of the data sample. Considering an input sequence $\mathbf{X}_t$ of all spike tokens from time chunk $t$ and a latent query $\mathbf{c}^{(t)}_{i}$ from learnable vector $i$, 
the encoder output is calculated via scaled dot-product attention \cite{vaswani_attention_2017}:
\begin{align*}
\mathbf{z}^{(t)}_{i}=\mathrm{softmax}\left(\frac{\mathbf{q}^{(t)}\mathbf{K}_t^\top}{\sqrt{D}}\right)\mathbf{V}_t,
\end{align*}
where $\mathbf{q}^{(t)} = \mathbf{c}^{(t)}_{i}\mathbf{W}_q$, $\mathbf{K}_t = \mathbf{X}_t\mathbf{W}_k$ and $\mathbf{V}_t = \mathbf{X}_t\mathbf{W}_v$ are the projected queries, keys, and values, respectively. The process is repeated for $T_c$ time chunks to obtain the latent token sequence $\mathbf{Z}=[\mathbf{z}^{(1)}_1,...,\mathbf{z}^{(1)}_{N_c},...,\mathbf{z}^{(T_c)}_1,...,\mathbf{z}^{(T_c)}_{N_c}]$.
\subsection{Backbone Architecture}
\paragraph{Attention-based.}
The attention-based backbone is composed of stacked self-attention blocks operating over the entire latent sequence \cite{azabou2023unified}. We denote $\mathbf{h}^{(t)}_{l,i}$ as the hidden state from latent query $i$ at time $t$ from layer $l$,
\begin{align*}
\mathbf{h}^{(t)}_{l,i}=\mathrm{softmax}\left(\frac{\mathbf{q}^{(t)}_l\mathbf{K}^\top}{\sqrt{D}}\right)\mathbf{V},
\end{align*}
where $\mathbf{q}^{(t)}_l = \mathbf{h}^{(t)}_{l-1,i}\mathbf{W}_q$ and $\mathbf{h}^{(t)}_{0,i} = \mathbf{z}^{(t)}_i$. 
The time complexity of this backbone is quadratic, but with respect to $N_z$ which is independent of the number of spike tokens. Consequently, the computation remains constant and computationally tractable even when handling very long input sequences \cite{jaegle2022perceiver}.
\paragraph{Recurrence-based.}
The recurrent backbone is a stack of SSM blocks \cite{ryoo2025generalizable} where the hidden state is updated following the transition function
\begin{align*}
\mathbf{h}^{(t)}_l = f_{\text{SSM}}(\mathbf{h}^{(t)}_{l-1}, \mathbf{h}^{(t-1)}_l).
\end{align*}
where $\mathbf{h}^{(t)}_0 = g( [\mathbf{z}^{(t)}_1,...,\mathbf{z}^{(t)}_{N_c}])$. Following \citet{ryoo2025generalizable}, we set $N_c=1$ and $g(\cdot)$ as an identity mapping.
\subsection{Masked Autoencoding}
The masked autoencoder uses a learnable mask token to mask out any of the latent tokens. After the masking step, we provide the latent tokens an additional learnable position embedding to encode the spatial relationships among latents, i.e., which latent query it originated from. We used a masking ratio of $50\%$, and as shown in Section \ref{app:sec:abl:maskratio}, our method is robust with respect to the specific value of the masking ratio. At the output, for each masked latent timestep, up to 10 neural units are randomly drawn uniformly without replacement, whose unit embeddings at that time are used as the queries for spike rate prediction.
\subsection{Output Cross-Attention for SL Pathways}
\label{app:sec:outputca}
For the SL pathway, each output query $\mathbf{o}_q$ attends to either all or the $k$ most recent latents depending on whether the backbone is \poyo or \possm, respectively. $\mathbf{q}_o$ is encoded by a learnable session embedding as well as the prediction timestamps. The output cross-attention facilitates flexible output predictions, enabling unaligned, irregular, and even ``out-of-boundary queries'' \cite{ryoo2025generalizable}.
\paragraph{Multi-task Supervised Learning.} The multi-task setting for the SL pathway is designed to be dataset-specific, by initiating separate linear readout layers for each of the tasks present in the dataset. Note that this configuration can be employed with both \poyo and \possm backbones, with the former named as \poyoplus \cite{azabou2025multisession} and latter as \possmplus. Note that in this paper, we enable the multi-task setting by default, and we will abuse the notation and refer to them as \poyo and \possm for brevity.
\subsection{Finetuning Strategy}
\label{app:sec:finetuning}
We adopt the two finetuning strategies from \citet{azabou2023unified} when adapting the pretrained model to previously unseen data sessions: Unit Identification (UI) and Full Finetuning (FT).
\paragraph{Unit Identification.} Under the UI strategy, we update only the unit and session embeddings during finetuning while keeping the rest of model frozen. This effectively transfers the pretrained model by mapping new unidentified units and sessions into the learned embedding space \cite{azabou2023unified}, providing an efficient strategy for rapid adaptation to new sessions. When transferring to sessions involving new tasks, UI additionally unfreezes task-specific embeddings and the linear readouts. 
\paragraph{Full Finetuning.} In contrast, FT unfreezes all model parameters during finetuning, therefore more closely resembling single-session end-to-end training. FT is typically performed in a staged manner: UI is applied first for a number of epochs, after which the remaining model parameters are unfrozen. Unless otherwise mentioned, UI was run for 100 epochs before unfreezing all parameters in our experiments. 
\subsection{Implemented Models on Monkey and Mouse Datasets}
Model hyperparameters of \mojo for both monkey and mouse datasets are shown in Table \ref{app:tab:hyperspikepoyo}, \ref{app:tab:hyperspikepossm}, \ref{app:tab:hyperspikepoyol}, and \ref{app:tab:hyperspikepoyoibl}, respectively. For \mojo-\pogru and \mojo-\pomamba, we adopted exactly the same model architecture reported in \citet{ryoo2025generalizable} with 20 ms chunks. For \mojo-\poyo, a slightly smaller model than \poyo-1 in \citet{azabou2023unified} was chosen to more closely match the parameters of the other \mojo backbones. We also adopted pathway integration to share the encoder computation between SSL and SL pathways in \mojo-\poyo, which resulted in 50 latent timesteps instead of the 8 used in \poyo-1. \mojo-\poyo-L is a larger model with parameter counts comparable to \poyo-1. For mouse decision tasks, we used a smaller \mojo-\poyo model. 
\vspace{5mm}
\begin{table}[H]
\centering
\begin{minipage}{0.45\textwidth}
\centering
\caption{\textbf{Hyperparameters for \mojo-\poyo.} Number of parameters counted for encoder, backbone and decoder.}
\vspace{0.1in}
\label{app:tab:hyperspikepoyo}
\begin{tabular}{lc}
\toprule
Hyperparameters & \#\\
\midrule
Input Dim & 128 \\
Backbone Layers & 20 \\
Latents per Chunk & 2 \\
Cross Heads & 1 \\
Self Heads & 8 \\
Params & 9.88M\\
\bottomrule
\end{tabular}
\end{minipage}
\hfill
\begin{minipage}{0.45\textwidth}
\centering
\caption{\textbf{Hyperparameters for \mojo-\poyo-L.} Number of parameters counted for encoder, backbone and decoder.}
\vspace{0.1in}
\label{app:tab:hyperspikepoyol}
\begin{tabular}{lc}
\toprule
Hyperparameters & \#\\
\midrule
Input Dim & 128 \\
Backbone Layers & 24 \\
Latents per Chunk & 4 \\
Cross Heads & 4 \\
Self Heads & 8 \\
Params & 12.06M\\
\bottomrule
\end{tabular}
\end{minipage}
\medskip
\end{table}

\vspace{5mm}
\begin{table}[H]
\centering
\begin{minipage}{0.45\textwidth}
\centering
\caption{\textbf{Hyperparameters for \mojo-\possm (Mamba).} Number of parameters counted for encoder, backbone and decoder.}
\vspace{0.1in}
\label{app:tab:hyperspikepossm}
\begin{tabular}{lc}
\toprule
Hyperparameters & \#\\
\midrule
Input Dim & 256 \\
Backbone Layers & 4 \\
Latents per Chunk & 1 \\
Cross Heads & 1 \\
RNN Hidden Dim & 512 \\
Params & 9.47M\\
\bottomrule
\end{tabular}
\end{minipage}
\hfill
\begin{minipage}{0.45\textwidth}
\centering
\caption{\textbf{Hyperparameters for \mojo-\poyo on Mouse Decision Tasks.} Number of parameters counted for encoder, backbone and decoder.}
\vspace{0.1in}
\label{app:tab:hyperspikepoyoibl}
\begin{tabular}{lc}
\toprule
Hyperparameters & \#\\
\midrule
Input Dim & 128 \\
Backbone Layers & 12 \\
Latents per Chunk & 4 \\
Cross Heads & 2 \\
Self Heads & 8 \\
Params & 6.27M\\
\bottomrule
\end{tabular}
\end{minipage}
\medskip
\end{table}

\vspace{5mm}
\begin{table}[H]
\centering
\begin{minipage}{0.45\textwidth}
\centering
\caption{\textbf{Hyperparameters for \mojo-\poyo on human ECoG dataset.} Number of parameters counted for encoder, backbone and decoder.}
\vspace{0.1in}
\label{app:tab:hyperspikepoyoecog}
\begin{tabular}{lc}
\toprule
Hyperparameters & \#\\
\midrule
Input Dim & 128 \\
Backbone Layers & 12 \\
Encoder Self-Attention Layers & 2 \\
Latents per Chunk & 8 \\
Cross Heads & 4 \\
Self Heads & 8 \\
Patch Size & 4 \\
Params & 7.44M\\
\bottomrule
\end{tabular}
\end{minipage}
\hfill
\begin{minipage}{0.45\textwidth}
\centering
\caption{\textbf{Hyperparameters for \poyo on human ECoG dataset.} Number of parameters counted for encoder, backbone and decoder.}
\vspace{0.1in}
\label{app:tab:hyperspikepoyo1ecog}
\begin{tabular}{lc}
\toprule
Hyperparameters & \#\\
\midrule
Input Dim & 128 \\
Backbone Layers & 6 \\
Encoder Self-Attention Layers & 2 \\
Latents per Chunk & 8 \\
Cross Heads & 4 \\
Self Heads & 8 \\
Patch Size & 4 \\
Params & 4.42M\\
\bottomrule
\end{tabular}
\end{minipage}
\medskip
\end{table}
\subsection{Implemented Models on Human Datasets}
We conducted hyperparameter tuning on \mojo and \poyo. Model hyperparameters of \mojo for human datasets are shown in Table \ref{app:tab:hyperspikepoyoecog}, and it needs a larger model mainly for better learning of SSL objectives. For \poyo, we found a larger model would sometimes fail the optimization, hence we sticked to a smaller model. A statistical baseline was implemented on this dataset, where independent component analysis (ICA) with 150 components was first run on the neural data, following which a linear support vector classifier was fitted on the ICA-transformed data. We used \texttt{fastICA} and \texttt{LinearSVC} from \texttt{scikit-learn} in our implementation. For EEGNet, we re-implemented the training loop with code and default hyperparameters from torcheeg packages \cite{zhang2024torcheeg}.
\section{Training Details}
\subsection{Monkey Reaching Tasks}
In each session from the monkey reaching datasets, the decoder is tasked with decoding 2-dimensional hand velocity from spiking data. For all experiments conducted on this task, we used mean squared error as the loss function. During training, behaviour signals up to 100 Hz were used as targets, and for sessions or datasets with a higher behaviour sampling rate, random subsampling was adopted to enforce the 100 Hz upper-bound.

We trained single-session and multi-session \poyo and \possm with a learning rate of 0.004 and 0.002 respectively using the Lamb \cite{lamb} optimizer. Batch sizes of 128 and 256 were used, respectively, an training was done for 500 epochs. Single-session training hyperparameters were adopted during UI and FT. 

Single-session \mojo were trained with a layer-wise learning rate of 0.01 on the unit embedding layer and 0.001 for the rest of the model using Lamb. Batch sizes were 128 and 64 for \mojo-\possm and \mojo-\poyo. For multi-session training, we implemented different learning rates for \mojo-\possm and \mojo-\poyo. For \mojo-\possm, the same layer-wise learning rate as single-session models was adopted, while for \mojo-\poyo, we used 2.5$e^{-4}$ for all parameters with AdamW \cite{adamw}. Batch sizes were 128 and 64 for \mojo-\possm and \mojo-\poyo, respectively. Note that we found that empirically, training \mojo-\possm with AdamW led to unstable optimization, and for Lamb, a larger learning rate was necessary for better learning on SSL related unit embeddings. During UI and FT, 0.01/0.001 with Lamb was used for \mojo-\possm and 5$e^{-4}$ with AdamW was used for \mojo-\poyo. All models were trained for 500 epochs. Multi-session training was done on 4 H100 GPUs, while single-session and finetuning were done on 1 NVIDIA L40S.

The NDT-2 \citep{ye_neural_2023}, NDT-3 \citep{ye2025a} and NEDS \citep{zhang2025neuralencodingdecodingscale} baselines were trained using the original code released by the authors, with some modifications to ensure that the models were being trained and evaluated on the exact same data splits as \mojo. We followed the original authors' suggestions and hyperparameter configurations for both training and finetuning these models on our datasets.

\subsection{Mouse Vision Tasks}
Each session of the mouse vision tasks presents a multi-task learning problem involving 2 or 3 (natural scenes only for sessions in stimulus set 1) multi-class classification tasks, and we used cross-entropy as the loss function for all of them. 

Separate baseline MLPs were trained for each individual task, where for drifting gratings orientation and temporal frequency (2 s per trial), we used 0.004 learning rate with AdamW, 128 batch size, 20 ms chunks, and a sequence length of 0.5 s. For natural scenes (250 ms per trial), a reduced learning rate of 0.0012 was used with sequence length of 250 ms. Learning rate reduction was necessary for stable training performance.

For multi-session training, we adopted task-specific weighting on different tasks following \citet{azabou2025multisession} for \poyo, where natural scene was assigned a weight of 0.3. However, we found that \pogru struggled to learn with the task weighting, and we instead adopted equal weights for all tasks. Note that the same applies to \mojo-\poyo and \mojo-\pogru. All other hyperparameters were kept the same as monkey tasks. 

For all single-session models, UI, and FT, equal weighting was adopted for all models. Hyperparameters were kept the same as monkey tasks for most models, except for single-session \mojo where larger learning rates were used (0.02/0.002), and for FT \mojo-\poyo where a smaller learning rate was used (2.5$e^{-4}$) due to overfitting.

The NEDS baseline was trained using the original code released by the authors, with several modifications to support the Allen visual coding Neuropixels dataset, non-simultaneous classification tasks, and ensure consistent data splits. We trained NEDS for 500 epochs on the Allen data. All other hyperparameters for training and finetuning followed the suggestions of the original authors.

\subsection{Mouse Decision Tasks}
We used cross-entropy loss for choice and block and mean squared error for wheel and whisker, and we weighted the loss of choice and block by $0.2$ since they were empirically found to be more prone to overfit.
For pretraining, AdamW optimizer was used with learning rate of $0.001$ and batch size of 64, while for finetuning learning rate of $1.25e^{-4}$ and batch size of 64. Pretraining was done on 2 NVIDIA L40S, while finetuning were done on 1 NVIDIA L40S with no UI for better stability across the 4 tasks. \poyo was pretrained and finetuned with Lamb optimizer with learning rate of $0.002$ and batch size of 64. NEDS was trained using the original code released by the author for this dataset. 
\subsection{Human Speech Tasks}
The tasks of classifying vowel, consonant, and consonant-vowel syllable from ECoG signals are each multi-class classification tasks with 3, 19, and 57 ground truth classes, and we used cross-entropy as the loss functions. ECoG is a new modality different from the original paper \citep{azabou2023unified,azabou2025multisession}, we used AdamW optimizer for both \mojo and \poyo and found it to be empirically good. For \poyo, the total number of syllables seen in the dataset was used as the output dimension of the readout layer across all three tasks. Hyperparameter tuning were run on \poyo on learning rate \{$2.5e^{-4}$,$5e^{-4}$\}, weight decay \{$1e^{-3}$,$1e^{-4}$\}, unit dropout (max/min/mode of units) \{256/64/128,128/32/64,64/16/32,32/8/16\}, and we also investigate different depth (6 or 12) and different output dimension (constant for all three tasks with total number of syllables or separate for each task based on its number of classes). As a result, \poyo were trained in 1 NVIDIA L40S, with $5e^{-4}$ learning rate with AdamW, weight decay of $1e^{-4}$, and unit dropout of 128/32/64. For \mojo, we adopted curriculum learning where we trained only SSL for 1000 epochs then proceed with the joint SSL-SL training. We found that warm start from a purely SSL checkpoint improved the performance of subsequent joint SSL-SL learning. In addition for \mojo, we tune the SSL loss coefficient in the joint loss function. \mojo was trained in 1 NVIDIA L40S, with $2.5e^{-4}$ learning rate with AdamW, weight decay of $1e^{-4}$, unit dropout of 256/64/128, and SSL loss coefficient of 0.5. 

Statistical methods were applied separately for each session on CPU. Three seeds were run for fitting the statistical models with random start but the variance was found to be trivial. EEGNet was run separately for each session and each task, with AdamW optimizer with learning rate of $0.001$ and batch size of 32. The Du-IN \cite{zheng2024duin} baselines were trained using the original code released by the authors, with some modifications to ensure that the models were being trained and evaluated on the exact same data splits as \mojo.
\section{Additional Experiments on Monkey Reaching Datasets}
In this section, we describe additional experiments conducted on monkey spiking datasets on various reaching tasks.
\subsection{Full Results on Reaching Tasks}
The complete results on the monkey reaching tasks are shown in Table \ref{app:tab:monkey20ms}, including single-session results as well as two additional \possm backbones: Mamba2 \cite{dao2024transformersssmsgeneralizedmodels} and xLSTM \cite{beck:24xlstm}. It can be seen from the table that certain backbones have better decoding performance across all evaluation sessions, namely transformer (as in \mojo-\poyo) and GRU (as in \mojo-\pogru). Based on our experience, transformer backbones usually exhibited better SSL performance during pretraining, due to the noncausal operations in self-attention, which in turns contributes to better performance in even an online evaluation strategy. Especially, transformer backbones demonstrated consistently better performance on the C-CO 2010 sessions, and this would relate to the fact that those sessions in general have more neural units. However, the inference speed of the transformer backbone still lag behind that of \possm, consistent with what was shown in \citet{ryoo2025generalizable}. Among the recurrent backbone, GRU stands out as the best performing architecture, especially in UI and single-session. For example, the single-session performance of \mojo-\pogru was on par with many pretrained model. One backbone choice worth highlighting is xLSTM, as it showed good balanced performance overall on both pretraining and finetuning.
\begin{table}[ht]
\centering
\vspace{5mm}
\caption{{\bf Behavioural decoding $R^2$ on monkey reaching tasks with 20 ms time chunks.} Values are mean $R^2$ $\pm$ SD over sessions. Best performing models are in boldface (1st) and underlined (2nd). Results marked by $^{\resultTaken}$ reproduced from \citet{ryoo2025generalizable}}
\vspace{0.1in}
\label{app:tab:monkey20ms}
{\renewcommand{\arraystretch}{1.2}
\adjustbox{max width=\textwidth}{\begin{NiceTabular}{c|>{\thinspace}l>{\enspace}c<{\enspace}>{\enspace}cc}
\toprule
 &  & \textit{Same Animal, Other days} & \multicolumn{2}{c}{\textit{New animal}} \\
 \cmidrule(lr){3-3} \cmidrule(lr){4-5} 
 & Method & C – CO 2010 (5) & T – CO (6) & T – RT (6) \\ \midrule
\parbox[t]{3mm}{\multirow{9}{*}{\rotatebox[origin=c]{90}{\textsc{\small From scratch}}}} & MLP$^{\resultTaken}$ & {0.5842 $\pm$ 0.2052} & 0.7940 $\pm$ 0.0341 & 0.6082 $\pm$ 0.3014 \\
 & Mamba$^{\resultTaken}$ & {0.6840 $\pm$ 0.0936} & 0.7318 $\pm$ 0.0426 & 0.6653 $\pm$ 0.0978 \\
 & GRU$^{\resultTaken}$ & 0.7742 $\pm$ 0.0964 & 0.8389 $\pm$ 0.0248 & 0.7414 $\pm$ 0.0426 \\
 & \poyo-SS & 0.7465 $\pm$ 0.1119 & 0.8509 $\pm$ 0.0423 & 0.6929 $\pm$ 0.0978 \\
 & \pomamba-SS$^{\resultTaken}$ & 0.7691 $\pm$ 0.0786 & 0.8613 $\pm$ 0.0121 & 0.7300 $\pm$ 0.0719 \\
 & \pogru-SS$^{\resultTaken}$ & \underline{0.7780} $\pm$ 0.0980 & 0.8724 $\pm$ 0.0190 & 0.7429 $\pm$ 0.0708 \\
 & \mojo-\poyo-SS & \textbf{0.8135} $\pm$ 0.0922 & \underline{0.8803} $\pm$ 0.0191 & \underline{0.7443} $\pm$ 0.0789 \\
 & \mojo-\pomamba-SS & 0.7752 $\pm$ 0.0878 & 0.8753 $\pm$ 0.0171 & 0.7428 $\pm$ 0.0861 \\
 & \mojo-\pogru-SS & 0.7768 $\pm$ 0.1283 & \textbf{0.8834} $\pm$ 0.0167 & \textbf{0.7711} $\pm$ 0.0596 \\
 \midrule
\parbox[t]{3mm}{\multirow{8}{*}{\rotatebox[origin=c]{90}{\textsc{\small Pretrained (UI)}}}}
 & \poyo (UI) & 0.7759 $\pm$ 0.1003 & 0.8123 $\pm$ 0.0419 & 0.7011 $\pm$ 0.0975 \\ 
 & \pomamba (UI) & 0.7283 $\pm$ 0.1138 & 0.8574 $\pm$ 0.0225 & 0.7283 $\pm$ 0.0846 \\
 & \pogru (UI) & 0.7632 $\pm$ 0.1013 & 0.8587 $\pm$ 0.0216 & 0.7331 $\pm$ 0.0775 \\
 & \mojo-\poyo (UI) & 0.7846 $\pm$ 0.0897 & 0.8387 $\pm$ 0.0251 & 0.7470 $\pm$ 0.0617 \\
 & \mojo-\pomamba (UI) & \underline{0.7937} $\pm$ 0.0735 & \underline{0.8753} $\pm$ 0.0160 & \underline{0.7591} $\pm$ 0.0588 \\
 & \mojo-\pogru (UI) & \textbf{0.7949} $\pm$ 0.0892 & \textbf{0.8772} $\pm$ 0.0189 & 0.7570 $\pm$ 0.0693\\
 & \mojo-\pomambatwo (UI) & 0.7720 $\pm$ 0.0872 & 0.8637 $\pm$ 0.0158 & 0.7565 $\pm$ 0.0582 \\
 & \mojo-\poxlstm (UI) & 0.7832 $\pm$ 0.1045 & 0.8671 $\pm$ 0.0255 & \textbf{0.7615} $\pm$ 0.0539 \\
 \midrule
 \parbox[t]{3mm}{\multirow{9}{*}{\rotatebox[origin=c]{90}{\textsc{\small Pretrained (FT)}}}}
 & NDT-2$^{\resultTaken}$ & 0.7846 $\pm$ 0.1167 & 0.7173 $\pm$ 0.0443 & 0.6323 $\pm$ 0.1339 \\
 & NDT-3 & 0.7524 $\pm$ 0.1322 & 0.8576 $\pm$ 0.0313 & 0.7066 $\pm$ 0.0980 \\
 & NEDS & 0.5968 $\pm$ 0.0760 & 0.7635 $\pm$ 0.0758 & 0.6121 $\pm$ 0.0918\\
 & \poyo (FT) & \underline{0.8244} $\pm$ 0.0753 & 0.8817 $\pm$ 0.0352 & 0.7624 $\pm$ 0.0815 \\
 & \pomamba (FT) & 0.8142 $\pm$ 0.0763 & 0.8949 $\pm$ 0.0152 & 0.7580 $\pm$ 0.0745 \\
 & \pogru (FT) & 0.8126 $\pm$ 0.0892 & 0.8936 $\pm$ 0.0212 & 0.7575 $\pm$ 0.0875 \\ 
 & \mojo-\poyo (FT) & \textbf{0.8438} $\pm$ 0.0888 & \textbf{0.9131} $\pm$ 0.0177 & \textbf{0.7964} $\pm$ 0.0663 \\
 & \mojo-\pomamba (FT) & 0.8153 $\pm$ 0.0925 & 0.9043 $\pm$ 0.0182 & 0.7675 $\pm$ 0.0806 \\
 & \mojo-\pogru (FT) & 0.8222 $\pm$ 0.0950 & \underline{0.9103} $\pm$ 0.0168 & 0.7776 $\pm$ 0.0710 \\
 & \mojo-\pomamba2 (FT) &  0.8189 $\pm$ 0.0918 & 0.9045 $\pm$ 0.0162 & 0.7623 $\pm$ 0.0731 \\
 & \mojo-\poxlstm (FT) & 0.8207 $\pm$ 0.0928 & 0.9082 $\pm$ 0.0185 & \underline{0.7712} $\pm$ 0.0670 \\
 \bottomrule
\end{NiceTabular}}}
\medskip
\end{table}
\subsection{Additional Results on Brain Region Classification}
Table \ref{app:tab:monkeyregionclf} includes the complete results on brain region classification with pretrained monkey embedding. \texttt{LogisticRegressionCV} from scikit-learn was adopted as the classifier, with 5-fold cross-validation. The process was repeated for 5 seeds on each classification task.
\begin{table}[ht]
\centering
\vspace{5mm}
\caption{{\bf Brain region classification accuracies on pretrained monkey reaching datasets.} M1: primary motor, PMd: dorsal premotor, S1: primary somatosensory. Values are mean accuracies $\pm$ SD over 5 seeds. Best performing models are in boldface (1st) and underlined (2nd).}
\vspace{0.1in}
\label{app:tab:monkeyregionclf}
{\renewcommand{\arraystretch}{1.2}
\adjustbox{max width=\textwidth}{\begin{NiceTabular}{l>{\enspace}c<{\enspace}>{\enspace}c<{\enspace}>{\enspace}c<{\enspace}>{\enspace}ccc<{\enspace}}
\toprule
 & \citet{perich_miller_2018_dataset} & \citet{Churchland2012} & \citet{odoherty2017nonhuman} & \multicolumn{3}{c}{\textit{Joint Monkey}} \\
 \cmidrule(lr){2-2} \cmidrule(lr){3-3} \cmidrule(lr){4-4} \cmidrule(lr){5-7} 
 Method & M1 vs. PMd & M1 vs. PMd & M1 vs. S1 & Multi-region & Multi-subject & M1 in Multi-subject \\ \midrule
 Chance & 50.00 & 50.00 & 50.00 & 33.33 & 16.67 & 16.67 \\
 \poyo & 64.23 $\pm$ 0.75 & 74.36 $\pm$ 1.81 & 81.17 $\pm$ 0.99 & 69.65 $\pm$ 0.45 & 49.67 $\pm$ 0.72 & 61.50 $\pm$ 0.51 \\
 \pomamba & 64.06 $\pm$ 0.64 & 74.78 $\pm$ 2.09 & 82.15 $\pm$ 0.83 & 69.54 $\pm$ 0.37 & 46.36 $\pm$ 0.24 & 60.74 $\pm$ 0.43 \\
 \pogru & 65.77 $\pm$ 0.90 & 71.70 $\pm$ 1.35 & 81.64 $\pm$ 0.70 & 69.51 $\pm$ 0.47 & 46.57 $\pm$ 0.55 & 60.30 $\pm$ 0.92 \\
 \mojo-\poyo & \textbf{81.08} $\pm$ 0.49 & \underline{98.17} $\pm$ 0.47 & \textbf{89.34} $\pm$ 0.46 & \textbf{78.24} $\pm$ 0.42 & \underline{72.69} $\pm$ 0.87 & \underline{84.51} $\pm$ 0.63 \\
 \mojo-\pomamba & 77.32 $\pm$ 0.51 & 95.30 $\pm$ 1.08 & 87.72 $\pm$ 0.60 & 73.91 $\pm$ 0.52 & 68.18 $\pm$ 0.55 & 78.91 $\pm$ 0.33 \\
 \mojo-\pogru & \underline{80.43} $\pm$ 0.40 & \textbf{98.54} $\pm$ 0.54 & \underline{88.40} $\pm$ 0.17 & \underline{76.88} $\pm$ 0.40 & \textbf{75.74} $\pm$ 0.43 & \textbf{85.28} $\pm$ 0.36 \\
 \bottomrule
\end{NiceTabular}}}
\medskip
\end{table}
\subsection{Ablation Study}
\subsubsection{Pathway Integration}
\label{app:sec:pathway}
We compared the performance of \mojo-\poyo with or without pathway integration. Note that without pathway integration, the input cross-attention of the SL pathway is exactly as \poyo (with 8 set of latents and 125 ms apart) separated from the SSL pathway. In both cases, we used the same number of latents per time steps ($N_c = 2$), so that all latent queries were shared between SSL and SL pathways. As shown in Table \ref{app:tab:pathwayIntegration}, pathway integration led to better FT performance and worse UI. The increases in FT performance is likely due to an unfied input cross-attention computation. Recall that even in the case of separate encoder computation, their parameters are still shared. Thus \mojo encoder would need to deal with discrepancy in latent time resolution, as latents in SSL pathway are 20 ms apart while the interval in SL pathway is 125 ms. On the other hand, separate encoder pathway could potentially ease the learning process for unit embeddings, since separate subsequent encoders can further process the unit embedding for SSL or SL with different focuses. The model with separate encoder also yields slightly worse inference speed, due to duplication in input cross-attention computations.
\begin{table}[H]
\centering
\vspace{5mm}
\caption{{\bf Ablation results on pathway integration for \mojo-\poyo.}}
\vspace{0.1in}
\label{app:tab:pathwayIntegration}
{\renewcommand{\arraystretch}{1.2}
\adjustbox{max width=\textwidth}{\begin{NiceTabular}{lccc}
\toprule
 Method & C – CO 2010 (5) & T – CO (6) & T – RT (6) \\ \midrule
 \mojo-\poyo (Shared Enc) (UI) & 0.7846 $\pm$ 0.0897 & 0.8387 $\pm$ 0.0251 & 0.7470 $\pm$ 0.0617 \\
 \mojo-\poyo (Sepa. Enc) (UI) & 0.7902 $\pm$ 0.0916 & 0.8523 $\pm$ 0.0200 & 0.7432 $\pm$ 0.0876 \\
 \mojo-\poyo (Shared Enc) (FT) & 0.8438 $\pm$ 0.0888 & 0.9131 $\pm$ 0.0177 & 0.7964 $\pm$ 0.0663 \\
 \mojo-\poyo (Sepa. Enc) (FT) & 0.8324 $\pm$ 0.1030 & 0.9030 $\pm$ 0.0164 & 0.7810 $\pm$ 0.0926 \\
 \bottomrule
\end{NiceTabular}}}
\medskip
\end{table}
\begin{table}[ht]
\centering
\vspace{2mm}
\caption{\textbf{Ablation results on latent masking strategy and ratio.} Results are shown for temporal and forward latent mask strategy on test $R^2$, SL test loss and SSL test loss for pretraining and test $R^2$ for finetuning. }
\medskip
{\renewcommand{\arraystretch}{1.2}
\adjustbox{max width=\textwidth}{\begin{NiceTabular}{c|>{\thinspace}c>{\enspace}ccc<{\enspace}>{\enspace}c}
\toprule
 & Mask Ratio & \multicolumn{3}{c}{Pretraining} & Finetuning \\
\cmidrule(lr){3-5} \cmidrule(lr){6-6} 
& & $R^2$ & MSE & Poisson NLL & $R^2$ \\ \midrule
\parbox[t]{3mm}{\multirow{7}{*}{\rotatebox[origin=c]{90}{\textsc{\small Temporal}}}} & 0.02 & 0.9166 & 0.01333 & 0.3848 & 0.8453 \\
& 0.05 & 0.9167 & 0.01303 & 0.3894 & 0.8447 \\
& 0.1 & 0.9178 & 0.01307 & 0.3861 & 0.8449 \\
& 0.3 & 0.9181 & 0.01295 & 0.3868 & 0.8442 \\
& 0.5 & 0.9178 & 0.01299 & 0.3871 & 0.8449 \\
& 0.7 & 0.9173 & 0.01324 & 0.3894 & 0.8430 \\
& 0.9 & 0.9167 & 0.01325 & 0.3964 & 0.8420 \\
\midrule
\parbox[t]{3mm}{\multirow{3}{*}{\rotatebox[origin=c]{90}{\textsc{\small Forward}}}} & 0.3 & 0.9161 & 0.01326 & 0.3874  & 0.8423 \\
& 0.5 & 0.9172 & 0.01303 & 0.3874 & 0.8467 \\
& 0.7 & 0.9158 & 0.01327 & 0.3915 & 0.8449 \\
\bottomrule
\end{NiceTabular}}}
\label{app:tab:maskratio}
\vspace{2mm}
\end{table}
\subsubsection{Masking Strategy and Ratio}
\label{app:sec:abl:maskratio}
Varying masking ratio from 0.02 to 0.9 for temporal masking (masking out percentage of latent time bins randomly) and from 0.3 to 0.7 for forward masking (masking out percentage of latent time bins in the future), we pretrained \mojo-\pogru on the entire Perich et al. dataset excluding the evaluation sessions and reported their finetuning performance on held-out monkey T sessions (Table \ref{app:tab:maskratio}). The results show that although masking ratio affects SSL training performance, the SL scores remain robust, likely due to the joint SSL-SL training strategy balancing out the 2 objectives automatically.
\subsubsection{Alternative SSL methods}
Comparison with alternative SSL methods is shown in Table \ref{app:tab:ssl}, where each model is pretrained on the entire Perich et al. dataset and finetuned on held-out monkey T sessions. Since \mojo performs masked autoencoder with temporal masking, as alternatives we consider joint SSL-SL training with three spatial masking schemes, where spikes from 1) randomly selected neurons in each bins; 2) randomly selected neuron across all bins; 3) randomly selected brain regions (M1 or PMd) are masked from input token sequences, as well as joint SSL-SL training with contrastive predictive coding (CPC) \cite{DBLP:journals/corr/abs-1807-03748} with two ways to choose negative samples: from the same sequence or from the different sessions. Note that finetuning was not conducted for input region mask and CPC with mixed session, as the former significantly under-performed during the pretraining and the latter by definition can not have negative samples during single-session finetuning. \mojo demonstrates improved performance over all alternative SSL methods on this dataset. In addition, we note that CPC same sequence showed severely overfitting on the contrastive loss, suggesting poorly integrated SSL-SL objectives in that case.
\vspace{5mm}
\begin{table}[ht]
\centering
\vspace{2mm}
\caption{\textbf{Ablation results on alternative SSL methods.} Results are shown for three input masking strategy and CPC with two negative samples range, on test $R^2$, SL test loss and p-value vs. \mojo-\pogru for pretraining and test $R^2$ for finetuning.}
\medskip
{\renewcommand{\arraystretch}{1.2}
\adjustbox{max width=\textwidth}{\begin{NiceTabular}{l>{\enspace}ccc<{\enspace}>{\enspace}c}
\toprule
 Model & \multicolumn{3}{c}{Pretraining} & Finetuning \\
\cmidrule(lr){2-4} \cmidrule(lr){5-5} 
& $R^2$ & MSE & p vs. \mojo-\pogru & $R^2$ \\ \midrule
 \pogru & 0.9107 & 0.01449 & < 0.001& 0.8319 \\
 \mojo-\pogru & 0.9178 & 0.01299 & N/A & 0.8449 \\
 Input Random Mask & 0.9080 & 0.01491 & \enspace0.003 & 0.8318 \\
 Input Neuron Mask & 0.9068 & 0.01502 & < 0.001 & 0.8322 \\
 Input Region Mask & 0.3419 & 0.12182 & < 0.001 & N/A\\
 CPC Same Sequence & 0.9114 & 0.01441 & < 0.001 & 0.8274\\
 CPC Mixed Session & 0.9060 & 0.01534 & < 0.001 & N/A\\
\bottomrule
\end{NiceTabular}}}
\label{app:tab:ssl}
\vspace{2mm}
\end{table}
\subsection{Transferring to Human Handwriting}
\setlength{\intextsep}{0.0pt}
\begin{table}[h]
\centering
\caption{\textbf{Human handwriting classification accuracies for \mojo-\pogru.}  
Values are mean accuracy $\pm$ SD over 3 seeds.}
\medskip
\scalebox{1.0}{
{\renewcommand{\arraystretch}{1.2}
\begin{NiceTabular}{lc}
\toprule
Method & Acc. (\%) $\uparrow$ \\ \midrule
From scratch & 95.34 $\pm$ 0.36 \\
Pretrained & 97.73 $\pm$ 0.41 \\
\bottomrule
\end{NiceTabular}}}
\label{app:tab:handwriting}
\vspace{2mm}
\end{table}
We conducted a transfer learning experiment from a pretrained monkey model to a human handwriting task. The viability of such transfer was previously demonstrated in \citet{ryoo2025generalizable}, and we wanted to verify that this results retained with \mojo. Briefly speaking, the human handwriting dataset from \citet{WillettHandwriting} includes 9 sessions of a human participant doing imagined single character writing. Spike counts were recorded from two 96-channel microelectrode arrays implanted in motor cortex, pre-binned at 10 ms. We tested the performance of both pretrained and from-scratch \mojo-\pogru, on decoding the intended characters. As shown in Table \ref{app:tab:handwriting}, pretrained \mojo models outperformed the ones trained from scratch, even with the same number of parameters, verifying the transfer capability of \mojo from monkey motor task to human handwriting.
\subsection{Additional Results on Reaching Phase Transfer in Finetuning}
\label{app:sec:phase}
Following \citet{azabou2023unified}, we defined 5 different sub-task phases in each of the CO sessions, where all 5 phases were included during training with different weights, but only \texttt{REACH} phase was considered during evaluation.
\begin{itemize}
    \item \texttt{HOLD}: in each valid trial, from ``\textit{target\_on\_time}'' to ``\textit{go\_cue\_time}''
    \item \texttt{REACH}: in each valid trial, from ``\textit{go\_cue\_time}'' to ``\textit{stop\_time}''
    \item \texttt{RETURN}: in each valid trial, from ``\textit{stop\_time}'' to trial end
    \item \texttt{INVALID}: duration of invalid trials
    \item \texttt{RANDOM}: all others except for outliers
\end{itemize}

\begin{table}[h]
\centering
\vspace{5mm}
\caption{{\bf Finetuning to a new session from pretrained \mojo-\pogru, with only \texttt{RANDOM} phase labels.} \mojo had access to all unlabelled data and only \texttt{RANDOM} labels. Values are mean $R^2$ $\pm$ SD over 5 seeds.}
\vspace{0.1in}
\label{app:tab:randomPhase}
{\renewcommand{\arraystretch}{1.2}
\adjustbox{max width=\textwidth}{\begin{NiceTabular}{cc<{\enspace}>{\enspace}cccccc}
\toprule
 \multicolumn{2}{c}{C-CO} & \multicolumn{6}{c}{T-CO} \\
 \cmidrule(lr){1-2} \cmidrule(lr){3-8} 
 C-10/13 & C-10/21 & T-08/19 & T-08/21 & T-08/23 & T-09/03 & T-09/05 & T-09/09 \\ \midrule
{0.53 $\pm$ 0.06} & {0.52 $\pm$ 0.09} & {0.29 $\pm$ 0.12} & {-0.008 $\pm$ 0.03} & {-0.07 $\pm$ 0.22} & {0.09 $\pm$ 0.12} & {0.16 $\pm$ 0.13} & {0.17 $\pm$ 0.17} \\
 \bottomrule
\end{NiceTabular}}}
\medskip
\label{app:tab:random}
\end{table}
\vspace{5mm}
To evaluate if SSL on unlabelled data can help in deducing labelled behaviour, we conducted a series of phase transfer experiments, where \mojo had access to the labels of only a subset of phases, when finetuning to a new session. In the first experiments, only the \texttt{RANDOM} phase labels were accessible (Table \ref{app:tab:randomPhase}). Note that since \texttt{RANDOM} only occupies a very small percentage of the total data (typically less than 1\%), this experiment was not performed on purely supervised method. 

Subsequently, we conducted a second experiment where labels from both \texttt{RANDOM} and \texttt{INVALID} phases were given to the models (Table \ref{app:tab:twophasesSL}), and a third one where both spikes and labels were only from \texttt{INVALID} phase (Table \ref{app:tab:invaludphasesBoth}). 
\begin{table}[H]
\centering
\vspace{5mm}
\begin{minipage}{0.45\textwidth}
\centering
\caption{\textbf{Comparison on phase transfer during finetuning with labels from two phases.} All spikes and only the \texttt{RANDOM} and \texttt{INVALID} phase labels are accessible by the models. Values are mean $R^2$ $\pm$ SD among sessions.}
\vspace{0.1in}
\label{app:tab:twophasesSL}
\begin{tabular}{lcc}
\toprule
Method & C-CO (2) & T-CO (5) \\ \midrule
\pogru & 0.798 $\pm$ 0.03 & 0.653 $\pm$ 0.08\\
\mojo-\pogru & 0.891 $\pm$ 0.01 & 0.811 $\pm$ 0.03\\
\bottomrule
\end{tabular}
\end{minipage}
\hfill
\begin{minipage}{0.45\textwidth}
\centering
\caption{\textbf{Comparison on phase transfer during finetuning with only data from \texttt{INVALID} phase.} Spikes and labels of only \texttt{INVALID} phase are accessible by the models. Values are mean $R^2$ $\pm$ SD among sessions.}
\vspace{0.1in}
\label{app:tab:invaludphasesBoth}
\begin{tabular}{lcc}
\toprule
Method & C-CO (2) & T-CO (6) \\ \midrule
\pogru & 0.752 $\pm$ 0.02 & 0.685 $\pm$ 0.07 \\
\mojo-\pogru & 0.752 $\pm$ 0.03 & 0.738 $\pm$ 0.06\\
\bottomrule
\end{tabular}
\end{minipage}
\medskip
\end{table}

The results show that \mojo can leverage additional unlabelled to improve the decoding performance of a phase never encountered during finetuning, especially when it is transferred to a new animal. 

Lastly, it is important to note that since the model have no access to important meta data such as monkey identity, session recoding date and lab during finetuning, the \texttt{RANDOM} phase labels, despite its small size, is still imperative to include that during training, as otherwise \mojo would struggle to decode beyond random even for the sessions of a seen animal. This is likely due to the fact that \mojo has been exposed to many potential neural-to-behaviour mapping during pretraining on heterogeneous data source, and that \mojo (to certain extend \poyo as well) keeps important information in session or unit embedding that bridges neural activity and behaviour variables in different sessions, which needs to be properly re-learned in a new session.
\subsection{Additional Results on Pretraining with Unlabelled Data}
\label{app:sec:unlabelledPretrain}
\begin{figure*}
    \centering
    \includegraphics[width=0.7\textwidth]{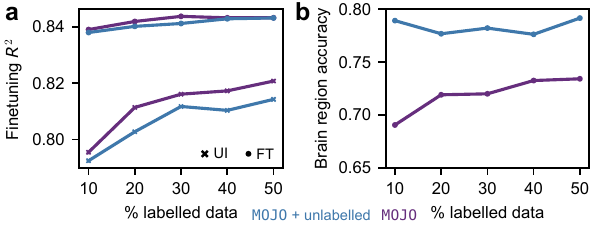}
    \caption{{\bf Additional results on pretraining with unlabelled data.} (a) Finetuning performance with additional unlabelled data during pretraining. (b) Brain region classification performance with additional unlabelled data during pretraining.}
    \label{fig:ftbrc}
\end{figure*}

For the models pretrained with varying ratio of unlabelled data, we tested their finetuning performance to new sessions (Figure \ref{fig:ftbrc}a), and we found that having additional unlabelled data during pretraining does not lead to better finetuning performance, especially in UI.  
Note that unlike the few-shot experiments described in section \ref{sec:monkeyexperiment}, no additional unlabelled data was used here during the finetuning. A plausible explanation is that the additional unlabelled data forces the model to prioritize on SSL objectives, enabling a better set of unit embeddings for pretraining datasets (Figure \ref{fig:ftbrc}b). But in the meanwhile, the SL performance becomes more dependent on the learned unit embedding, eventually harming the generalization performance. This result further validates the importance of labelled data during pretraining, as shown by upward trend in UI performance.

\subsection{Comparison on Running Time}
\setlength{\intextsep}{0.0pt}
\begin{table}[h]
\centering
\caption{\textbf{Per epoch running time comparison on monkey reaching datasets.}}
\medskip
\scalebox{1.0}{
{\renewcommand{\arraystretch}{1.2}
\begin{NiceTabular}{lc}
\toprule
Models & Epoch time (s) \\ \midrule
\pogru & 220 \\
\mojo-\pogru & 246 \\
\midrule
\poyo & 522 \\
\mojo-\poyo & 320 \\
\mojo-\poyo-L & 476 \\
\bottomrule
\end{NiceTabular}}}
\label{app:tab:time}
\vspace{2mm}
\end{table}
We compare running time for \mojo, \pogru and \poyo on monkey reaching datasets in Table \ref{app:tab:time}, where the reported numbers are obtained from running a full training epoch in 4 H100 GPUs. As shown in the first two rows on \possm backbone, \mojo does not incur large compute overheads, likely due to pathway integration strategy (see Section \ref{sec:integration}) between SSL and SL pathways. Note that numbers on \poyo backbone are not directly comparable due to a smaller backbone (\mojo-\poyo) and/or a shorter latent sequence (\mojo-\poyo and \mojo-\poyo-L). However, we choose to include it for the sake of completeness. 
\section{Additional Experiments on Mouse Vision Datasets}
In this section, we describe additional experiments conducted on mouse spiking datasets on various vision tasks.
\subsection{Full Results on Mouse Vision Tasks}
The complete results on the mouse vision tasks are shown in Table \ref{app:tab:mouse20ms}, including single-session results as well as all the tasks in 2 stimulus sets, including natural scenes (NS) classification of 119 classes, drifting grating orientation (DG) classification of 8 classes, and drifting grating temporal frequency (TF) classification of 5 classes. Note that in stimulus set 2, both DG and TF have reduced number of classes, including 4 orientations and 1 temporal frequency, hence we skip the TF results in stimulus 2 as it is trivial. Unlike the monkey results, single-session performance of \mojo underperformed those of purely supervised model, except for \mojo-\pogru in DG. We hypothesize that this discrepancy across datasets can be largely caused by the sheer difference in the number of units per session. Note that while the average number of units in monkey sessions is 57 for T and 191 for C-CO 2010, mice session has an average number of 676 units among evaluation sessions. The greatly increased number of units, and the more diverse brain regions, could make SSL difficult to learn, as \mojo needs to learn individual unit embeddings from scratch. The poorer performance in single-session \mojo is in direct contrast to pretrained model, where \mojo consistently outperformed SL methods in every tasks, reaching almost 100 \% accuracy on both DG and TF. 
\begin{table}[ht]
\centering
\vspace{5mm}
\caption{{\bf Visual stimuli classification accuracies on mouse vision tasks with 20 ms time chunks.} NS: Natural Scenes; DG: Drifting Grating Orientation; TF: Drifting Grafting Temporal Frequency. (A): with additional unlabelled data from Allen datasets. (J): with joint monkey-mouse dataset. Values are mean accuracies $\pm$ SD over sessions. Best performing models are in boldface (1st) and underlined (2nd), except for DG of stimulus set 2.}
\vspace{0.1in}
\label{app:tab:mouse20ms}
{\renewcommand{\arraystretch}{1.2}
\adjustbox{max width=\textwidth}
{\begin{NiceTabular}{c|>{\thinspace}l>{\enspace}ccc<{\enspace}c}
\toprule
 &  & \multicolumn{3}{c}{\textit{Stimulus Set 1}} & \textit{Stimulus Set 2} \\
 \cmidrule(lr){3-5} \cmidrule(lr){6-6}
 & Method & NS (4) & DG (4) & TF (4) & DG (4) \\ \midrule
\parbox[t]{3mm}{\multirow{5}{*}{\rotatebox[origin=c]{90}{\textsc{\small From scratch}}}} & Chance & 0.84 & 12.5 & 20 & 25 \\
& MLP & 83.17 $\pm$ 4.92 & 90.18 $\pm$ 2.03 & 87.11 $\pm$ 2.61 & 99.09 $\pm$ 0.63 \\
 & \poyo-SS & \underline{90.04} $\pm$ 3.92 & \underline{98.33} $\pm$ 2.10 & \underline{97.29} $\pm$ 5.00 & \underline{99.79} $\pm$ 0.24 \\
 & \pogru-SS & \textbf{91.11} $\pm$ 3.36 & 91.72 $\pm$ 3.29 &  \textbf{97.60} $\pm$ 2.36 & \textbf{99.95} $\pm$ 0.10 \\
 & \mojo-\poyo-SS & 89.08 $\pm$ 3.42 & 95.73 $\pm$ 3.40 &  94.27 $\pm$ 2.99 & 99.74 $\pm$ 0.10 \\
 & \mojo-\pogru-SS & 87.48 $\pm$ 5.26 & \textbf{98.39} $\pm$ 1.99 &  96.51 $\pm$ 5.61 & 99.69 $\pm$ 0.27 \\
 \midrule
\parbox[t]{3mm}{\multirow{6}{*}{\rotatebox[origin=c]{90}{\textsc{\small Pretrained (UI)}}}}
 & \poyo (UI) & 88.17 $\pm$ 3.17 & 93.54 $\pm$ 2.06 & 98.70 $\pm$ 5.98 & \underline{99.90} $\pm$ 0.12 \\ 
 & \pogru (UI) & 88.87 $\pm$ 2.55 & 99.17 $\pm$ 0.76 & 98.54 $\pm$ 0.85 & \textbf{99.95} $\pm$ 0.10 \\
 & \mojo-\poyo (UI) & 91.53 $\pm$ 1.99 & \underline{99.74} $\pm$ 0.31 & \underline{99.69} $\pm$ 0.36 & \textbf{99.95} $\pm$ 0.10 \\
 & \mojo-\pogru (UI) & 92.54 $\pm$ 3.41 & 99.53 $\pm$ 0.20 & 99.01 $\pm$ 0.55 & \textbf{99.95} $\pm$ 0.10 \\
 & \mojo-\poyo(A) (UI) & 92.06 $\pm$ 2.62 & \textbf{99.84} $\pm$ 0.31 & 99.64 $\pm$ 0.36 & 99.84 $\pm$ 0.20 \\
 & \mojo-\poyo(J) (UI) & \underline{93.15} $\pm$ 2.39 & 99.43 $\pm$ 1.15 & 99.43 $\pm$ 1.01 & \textbf{99.95} $\pm$ 0.10 \\
 & \mojo-\poyo-L(J) (UI) & \textbf{94.18} $\pm$ 2.00 & \textbf{99.84} $\pm$ 0.20 & \textbf{99.79} $\pm$ 0.29 & 99.95 $\pm$ 0.10\\
 \midrule
 \parbox[t]{3mm}{\multirow{6}{*}{\rotatebox[origin=c]{90}{\textsc{\small Pretrained (FT)}}}}
 & \poyo (FT) & 91.03 $\pm$ 4.07 & 93.39 $\pm$ 4.50 & 98.54 $\pm$ 0.66 & 99.69 $\pm$ 0.21 \\
 & \pogru (FT) & 91.09 $\pm$ 5.64 & 99.48 $\pm$ 0.62 & 98.85 $\pm$ 1.23 & \textbf{99.95} $\pm$ 0.10 \\ 
 & \mojo-\poyo (FT) & 94.12 $\pm$ 2.79 & \textbf{100.00} $\pm$ 0.0 & \underline{99.69} $\pm$ 0.40 & \textbf{99.95} $\pm$ 0.10 \\
 & \mojo-\pogru (FT) & 94.47 $\pm$ 2.83 & 99.22 $\pm$ 0.91 & 99.17 $\pm$ 1.01 & 99.74 $\pm$ 0.40 \\
 & \mojo-\poyo(A) (FT) & 94.05 $\pm$ 2.51 & \textbf{100.00} $\pm$ 0.0 & 99.27 $\pm$ 0.52 & \textbf{99.95} $\pm$ 0.10 \\
 & \mojo-\poyo(J) (FT) & \underline{94.62} $\pm$ 2.74 & 99.69 $\pm$ 0.64 & 99.38 $\pm$ 0.66 & \underline{99.79} $\pm$ 0.29 \\
 & \mojo-\poyo-L(J) (FT) & \textbf{95.48} $\pm$ 2.05 & \underline{99.74} $\pm$ 0.52 & \textbf{99.74} $\pm$ 0.31 & \textbf{99.95} $\pm$ 0.10\\
 \bottomrule
\end{NiceTabular}}}
\medskip
\end{table}
\subsection{Additional Results on Brain Region Classification}
\label{app:sec:mouseprobing}
\begin{figure}[t]
  \vskip 0.2in
    \includegraphics[width=\textwidth]{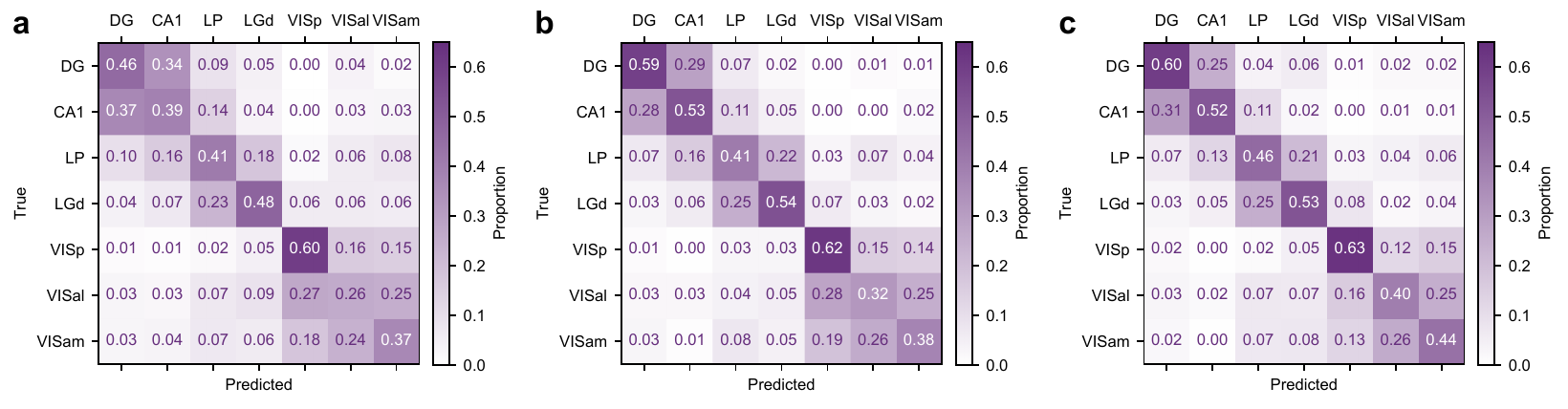}
    \caption{
      \textbf{Confusion matrix for mouse brain region classifications across different models.} Left: from model with paired data from Allen datasets; Middle: Left + additional unlabelled data from Allen datasets; Right: Middle + monkey dataset.
    }
    \label{fig:mouseregionclfthree}
\end{figure}
\begin{table}
\centering
\vspace{5mm}
\caption{{\bf Brain region classification accuracies on pretrained mouse vision datasets.} Values are mean accuracies $\pm$ SD over 5 seeds. (A): with additional unlabelled data from Allen datasets. (J): with joint monkey-mouse dataset. Best performing models are in boldface (1st) and underlined (2nd).}
\medskip
\scalebox{1.0}{
{\renewcommand{\arraystretch}{1.2}
\begin{NiceTabular}{lcc}
 \toprule
 Method & 3 merged-regions & 7 regions \\ 
 \midrule
 Chance & 33.33 & 14.29 \\
 ISI Baseline \citep{schneider2022} & 68.46 & 38.46 \\
 \pogru & 56.66 $\pm$ 0.78 & 26.93 $\pm$ 0.85 \\
 \poyo & 58.45 $\pm$ 0.79 & 27.99 $\pm$ 0.39 \\
 \mojo-\pogru & 70.09 $\pm$ 0.58 & 36.77 $\pm$ 0.86 \\
 \mojo-\poyo & 74.48 $\pm$ 0.64 & 43.17 $\pm$ 1.35 \\
 \mojo-\poyo(A) & 79.29 $\pm$ 0.65 & 48.67 $\pm$ 0.72 \\
 \mojo-\poyo(J) & \underline{80.41} $\pm$ 0.83 & \underline{51.16} $\pm$ 1.03 \\
 \mojo-\poyo-L(J) & \textbf{82.72} $\pm$ 0.40 & \textbf{54.78} $\pm$ 1.50 \\
 \bottomrule
\end{NiceTabular}}}
\label{app:tab:mousebrainregionclf}
\vspace{5mm}
\end{table}
We used logistic regression with 5-fold cross-validation as the classifier, repeated across 5 seeds. Notably, in this dataset the unit counts vary substantially across brain regions. Therefore to ensure balanced classes, we performed region-wise sub-sampling on the neural units prior to fitting each classifier.

Figure \ref{fig:mouseregionclfthree} plots confusion matrices on 7 region classification from several \mojo-\poyo model with different pretraining datasets. This includes (a) only labelled data, (b) with additional unlabelled data from the same dataset (\mojo-\poyo(A)), and (c) with additional monkey data (\mojo-\poyo(J)). Visual inspection reflects that the diagonal elements display larger weights as more and more unlabelled data is brought into the training pipeline. This visual observation also coincides with the quantitative results (Table \ref{app:tab:mousebrainregionclf}), where an increasing test accuracies on the 7 brain region classification can be observed with more unlabelled data. Note that we also compare with non-\mojo baselines, including pure SL \poyo as well as an ISI feature-based baseline \citep{schneider2022}, which computes 18 hand-crafted features for each neuron (ISI distributions, gamma-shape fits, band-limited PSDs of binned spike trains) from the entire dataset (not just the training set). \mojo-\poyo variants outperform all these baseline approaches.
\subsection{Additional Results on Neuronal Feature Prediction}
\label{app:sec:unitmetadata}

\begin{figure}[ht]
    \centering
    \includegraphics[width=\linewidth]{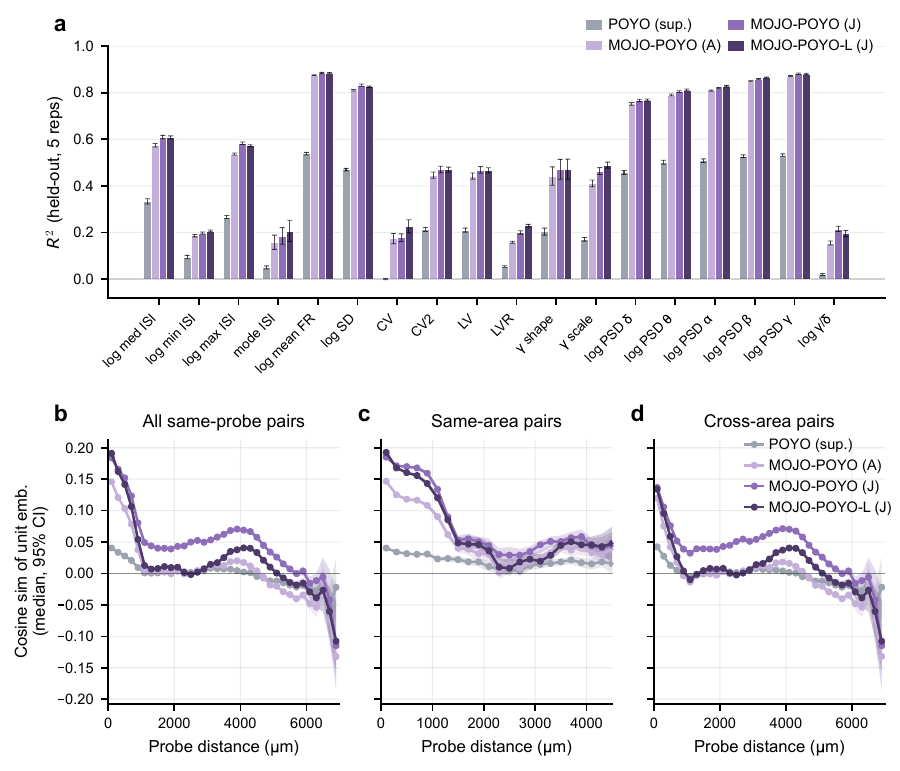}
    \caption{{\bf Unit-level metadata is encoded better by \mojo than by pure-SL \poyo.} (a) Linear-probe regression of 18 single-neuron spike-statistic features (ISI moments, firing rate, variability indices, gamma-distribution fit parameters, and band-limited PSDs of the binned spike train) from each model's unit embeddings on the Allen visual coding Neuropixels dataset. Bars show mean $R^2$ over 5 folds; error bars indicate s.d. across splits. (b--d) Cosine similarity between unit-embedding pairs as a function of probe distance for (b) all same-probe pairs (1.23M pairs), (c) same-area pairs (470k pairs, capped at 4.5\,mm where pair counts become sparse), and (d) cross-area pairs. Probe distance is measured along the Neuropixels shank as $|\Delta\text{channel}| \times 20\,\mu$m. Lines show the per-bin median cosine similarity (200\,$\mu$m bins); shaded bands are 95\% non-parametric CIs on the median. The mid-distance bump in cross-area pairs (panel d) is dominated by visual-cortex $\leftrightarrow$ visual subcortical pairs -- specifically thalamic (LGd, LP; $\sim$40\% of cross-area pairs in the 3.5--4.5\,mm range) and midbrain (APN; $\sim$14\%) partners of visual-cortical units (VISp, VISal, VISam, VISpm, VISrl). This indicates that MOJO embeddings cluster functionally related neurons across large anatomical separations -- a structure largely absent in the supervised baseline. All pairwise $\Delta$ Spearman $r$ differences between checkpoints are significant (paired bootstrap, $B=200$, $n_\text{resample}=3\times10^5$, all 95\% CIs exclude zero, $p<10^{-300}$).}
    \label{fig:unitmetadata}
\end{figure}

Given \mojo's strong performance in predicting metadata such as brain
regions from unit embeddings, we wanted to test more thoroughly     
whether the SSL objective allows the unit embeddings to encode       
finer-grained, neuron-level properties. We evaluated this on 50 Allen
visual coding Neuropixels sessions (34,606 single units across 7
target regions) along two complementary axes (Figure
\ref{fig:unitmetadata}). First, we used the unit embeddings to
regress 18 handcrafted single-neuron spike-statistic
features including moments of the ISI distribution, gamma-shape fits,
and band-limited PSD of the binned spike train \citep{schneider2022}. \mojo embeddings
predict these targets much better than pure-SL \poyo models across
every target and every region: e.g., for log(median ISI) $R^2$=0.57--0.61
(\mojo variants) vs 0.33 (\poyo); for log(mean firing rate) 0.88 vs
0.54; for the coefficient of variation, where supervised \poyo
decoding fails entirely, $\sim$0.20 vs $\sim$0.00. Second, we asked whether the
embedding geometry reflects probe topology by computing the cosine
similarity between pairs of unit embeddings on the same probe (1.23M
same-probe pairs). \mojo embeddings show a substantially stronger
anti-correlation between embedding similarity and electrode distance
than supervised \poyo (Spearman $r$ = -0.25 to -0.29 vs -0.12; all
pairwise $\Delta r$ CIs exclude zero, bootstrap $p < 1e-300$). Splitting these
pairs by area-pair type further reveals that, beyond purely
anatomical decay, \mojo embeddings cluster pairs of units across
distant but functionally related regions (e.g., cortex-thalamus at $\sim$4
mm probe distance, or visual cortex and visual-recipient subcortical regions) -- structure that is largely absent in the
supervised baseline.

\subsection{Additional Results on Joint Pretraining combining Monkey and Mouse Datasets}
\label{app:sec:transfer}
\begin{table}
\centering
\vspace{5mm}
\caption{{\bf Mouse Validation Decoding Accuracy at Epoch 199.} Best performing models are in boldface (1st) and underlined (2nd).}
\medskip
{\renewcommand{\arraystretch}{1.2}
\adjustbox{max width=\textwidth}
{\begin{NiceTabular}{c|>{\thinspace}l>{\enspace}cccc}
 \toprule
  & & \multicolumn{3}{c}{\textit{Validation Acc. ($\%$)}}\\
 \cmidrule(lr){3-6}
 & Method & NS & TF & DG & Avg. \\ 
 \midrule
 \parbox[t]{3mm}{\multirow{4}{*}{\rotatebox[origin=c]{90}{\textsc{\small Mouse}}}} & \poyo(Lamb) & 71.33 & 98.36 & 94.55 & 88.08 \\
 & \poyo(Adam) & 51.08 & 91.59 & 94.55 & 79.07 \\
 & \mojo-\poyo & \underline{73.01} & \underline{99.23} & \underline{98.02} & \underline{90.09} \\
 & \mojo-\poyo(A) & \textbf{74.47} & \textbf{99.49} & \textbf{98.82} & \textbf{90.93} \\
  \midrule
 \parbox[t]{3mm}{\multirow{4}{*}{\rotatebox[origin=c]{90}{\textsc{\small Joint}}}} & \poyo(Lamb)(J) & 65.72 & 97.01 & 95.29 & 86.00 \\
 & \poyo(Adam)(J)& 77.04 & 98.52 & 96.25 & 90.91 \\
 & \mojo-\poyo(J) & \underline{78.06} & \underline{99.00} & \textbf{98.20} & \underline{91.76} \\
 & \mojo-\poyo-L(J) & \textbf{79.50} & \textbf{99.21} & \underline{97.86} & \textbf{92.19} \\ 
 \bottomrule
\end{NiceTabular}}}
\label{app:tab:valjoint}
\vspace{5mm}
\end{table}
As shown in Table \ref{app:tab:valjoint}, for \mojo on mouse vision tasks, positive transfers via convergence speed can be observed on validation performance at early epochs, compared to mouse-only model. The joint model is also capable of retaining knowledge of the previously learned tasks, verified by its brain region classification and finetuning performance, which stayed consistent as the monkey-only model: $0.8551$ decoding $R^2$ averaged over monkey T sessions and $77.51\%$ multi-region classification from unit embeddings of pretrained monkey sessions, compared to $0.8548$ and $78.24\%$ for monkey-only model, respectively. With a larger model and joint training, \mojo improved further on monkey brain region classification to $81.02\%$, with similar finetuning performance of $0.8533$. 

The same curriculum learning experiment was also performed on \poyo model, and its results differed according to the choice of optimizer: With Lamb optimizer, no clear transfer was observed on validation curve, whereas with AdamW optimizer, the mouse-only model learned much more slowly, but the joint model was able to learn the mouse tasks much faster. Empirically, we found that AdamW optimizer induces larger changes on \poyo's unit embeddings, for example, the standard deviation of unit embedding at epoch $199$ of the joint training is $0.093$ for AdamW optimizer and $0.036$ for Lamb, from $0.053$ and $0.031$ at epoch $0$, respectively. This may be due to the additional per-layer rescaling introduced in Lamb optimizer \cite{lamb}. We can speculate that for training on more homogeneous dataset, such small changes on unit embedding might not be problematic, as the rest of the model can be trained to capture the variations in the dataset, but when such a model is subsequently transferred to a highly heterogeneous dataset, like cross-species and cross-task data in our case, larger embedding updates may facilitate transferring. Nevertheless, more comprehensive experiments are required to fully understand the root cause.
\section{Additional Experiments on Mouse Decision Datasets}
\label{app:sec:iblneds}
In this section, we include results of NEDS \cite{zhang2025neuralencodingdecodingscale}, rerun on mouse decision datasets due to changes on split and spike-sorting. The dataset changes are confirmed by the original authors. We ran NEDS before and after the most recent fix on masking in the official repository, and the results are shown in Table \ref{app:tab:nedsibl}.

\begin{table}
\centering
\vspace{5mm}
\caption{{\bf Decoding performance of NEDS on all behaviour tasks of mouse decision datasets.} Values are mean $R^2$ or balanced accuracies $\pm$ SD over sessions.}
\medskip
\scalebox{1.0}{
{\renewcommand{\arraystretch}{1.2}
\begin{NiceTabular}{lcccc}
 \toprule
 Method & Choice & Block & Wheel & Whisker \\ 
 \midrule
 NEDS & 0.8200 $\pm$ 0.1206 & 0.8116 $\pm$ 0.0876 & 0.5629 $\pm$ 0.0786 & 0.5066 $\pm$ 0.0975 \\
 NEDS+\href{https://github.com/yzhang511/NEDS/commit/61d2ef65556dbbc2b828d542f19bd4e7134c92f3}{bugfix} & 0.8202 $\pm$ 0.1045 & 0.7794 $\pm$ 0.1039 & 0.5152 $\pm$ 0.0902 & 0.4751 $\pm$ 0.1068 \\
 \bottomrule
\end{NiceTabular}}}
\label{app:tab:nedsibl}
\vspace{5mm}
\end{table}

\section*{Broader Impact}
The method proposed in this paper could contribute to the advancement of brain-computer interfaces to restore motor and speech functions. Datasets used to train the proposed approach originate from animal or human experiments involving surgical implanting, raising ethical concerns common to many medical technologies, including, but not limited to, privacy protection, harm reduction and animal welfare. It is therefore imperative to ensure all experimental protocols comply with IRB guidelines. 
In addition, the research described in this paper could potentially be deployed in real-world settings, which may lead to negative financial and social impacts. Therefore, rigorous testing and comprehensive evaluation are required before any practical deployment.

\clearpage

\end{document}